\PassOptionsToPackage{unicode}{hyperref}
\PassOptionsToPackage{hyphens}{url}
\PassOptionsToPackage{dvipsnames,svgnames,x11names}{xcolor}
\documentclass[
  10pt,
]{article}
\usepackage{amsmath,amssymb}
\usepackage{iftex}
\ifPDFTeX
  \usepackage[T1]{fontenc}
  \usepackage[utf8]{inputenc}
  \usepackage{textcomp} 
\else 
  \usepackage{unicode-math} 
  \defaultfontfeatures{Scale=MatchLowercase}
  \defaultfontfeatures[\rmfamily]{Ligatures=TeX,Scale=1}
\fi
\usepackage{lmodern}
\ifPDFTeX\else
\fi
\IfFileExists{upquote.sty}{\usepackage{upquote}}{}
\IfFileExists{microtype.sty}{
  \usepackage[]{microtype}
  \UseMicrotypeSet[protrusion]{basicmath} 
}{}
\makeatletter
\@ifundefined{KOMAClassName}{
  \IfFileExists{parskip.sty}{%
    \usepackage{parskip}
  }{
    \setlength{\parindent}{0pt}
    \setlength{\parskip}{6pt plus 2pt minus 1pt}}
}{
  \KOMAoptions{parskip=half}}
\makeatother
\usepackage{xcolor}
\usepackage[margin=1in]{geometry}
\usepackage{longtable,booktabs,array}
\usepackage{calc} 
\usepackage{etoolbox}
\makeatletter
\patchcmd\longtable{\par}{\if@noskipsec\mbox{}\fi\par}{}{}
\makeatother
\IfFileExists{footnotehyper.sty}{\usepackage{footnotehyper}}{\usepackage{footnote}}
\makesavenoteenv{longtable}
\usepackage{graphicx}
\makeatletter
\def\maxwidth{\ifdim\Gin@nat@width>\linewidth\linewidth\else\Gin@nat@width\fi}
\def\maxheight{\ifdim\Gin@nat@height>\textheight\textheight\else\Gin@nat@height\fi}
\makeatother
\setkeys{Gin}{width=\maxwidth,height=\maxheight,keepaspectratio}
\makeatletter
\def\fps@figure{htbp}
\makeatother
\providecommand{\tightlist}{%
  \setlength{\itemsep}{0pt}\setlength{\parskip}{0pt}}
\ifLuaTeX
\usepackage[bidi=basic]{babel}
\else
\usepackage[bidi=default]{babel}
\fi
\babelprovide[main,import]{american}

\def\languageshorthands#1{}
\ifLuaTeX
  \usepackage{selnolig}  
\fi
\usepackage{bookmark}
\IfFileExists{xurl.sty}{\usepackage{xurl}}{} 
\hypersetup{
  pdftitle={QuantumPhaseNet: A Gauge-Covariant Geometric and Quantum-Spectral Theory of Semantic Concept Hierarchies with Prototype Validation of a Classical Quantum-Inspired Model},
  pdfauthor={Kiyotaka Kasubuchi; Kazuo Fukiya},
  pdflang={en-US},
  colorlinks=true,
  linkcolor={blue},
  filecolor={Maroon},
  citecolor={Blue},
  urlcolor={blue},
  pdfcreator={LaTeX via pandoc}}

\title{QuantumPhaseNet: A Gauge-Covariant Geometric and Quantum-Spectral
Theory of Semantic Concept Hierarchies with Prototype Validation of a
Classical Quantum-Inspired Model}
\author{Kiyotaka Kasubuchi \and Kazuo Fukiya}
\date{July 30, 2026}

\begin{document}
\maketitle
\begin{abstract}
We present QuantumPhaseNet, a gauge-covariant geometric and
quantum-spectral extension of Transformer representations.
Context-dependent semantic states are modeled as complex amplitudes; a
covariant phase rate induces a semantic wavelength used as a proxy for
conceptual scale; and low-frequency graph modes define a document-level
discourse direction. The theoretical part establishes local gauge
invariance, unitarity of the quantum block, boundedness and conditional
stability of WavePhase Attention, and a calibratable hallucination-risk
formulation. We also implemented a fully offline Validation Studio for
the classical quantum-inspired pipeline in Section 14.1 and evaluated
the five research questions in Section 16.1 on its built-in synthetic
setting (n=240, observation noise 0.22, circuit noise 0.08, five seeds).
RQ1 yielded a wavelength-hierarchy Spearman correlation of 0.852 versus
0.707 for the baseline, 87.3\% direction accuracy, and AUC 0.953. RQ2
achieved discourse alignment 0.933 versus 0.589 and 41.2 versus 16.2
paragraphs before drift. RQ3 achieved AUROC 0.881 versus cosine 0.765
and phase-shuffle 0.536. RQ4 achieved error-detection AUROC 0.854 versus
entropy 0.634, with Brier 0.150 and ECE 0.098. RQ5 did not show quantum
advantage: target probability and end-to-end cost efficiency were 25.5\%
and 0.107, compared with 70.7\% and 0.707 for the Chebyshev classical
approximation. These results provide initial synthetic evidence for the
classical quantum-inspired components, but not external validity or
unconditional quantum speedup.
\end{abstract}


\section{1. Introduction}\label{introduction}

\subsection{1.1 Background}\label{background}

Transformers compute dependencies within a sequence in parallel through
self-attention. For an input representation
\(X \in \mathbb{R}^{n \times d}\), standard self-attention is

\[Q = XW_{Q},\quad\quad K = XW_{K},\quad\quad V = XW_{V},\]

\[\operatorname{Att}(Q,K,V)=\operatorname{softmax}\left(\frac{QK^{\top}}{\sqrt{d_k}}+M_{\mathrm{causal}}\right)V\]

as introduced by Vaswani et al.~(2017). Although this architecture is
highly expressive, inner-product similarity alone does not explicitly
reveal concept inclusion, levels of abstraction, parallel transport
between contexts, document-level discourse direction, connection to
evidence, or the geometry of a generation trajectory.

Order Embeddings learn semantic hierarchies as explicit coordinate-wise
orders (Vendrov et al.~2016), while Poincaré Embeddings exploit the
correspondence between tree-like volume growth and negative curvature
(Nickel and Kiela 2017). Geometric deep learning incorporates symmetry
and geometry into learning systems (Bronstein et al.~2021); graph
Fourier analysis provides spectral tools on irregular domains (Shuman,
Ricaud, and Vandergheynst 2013); and vector diffusion based on the
connection Laplacian compares vectors that live in different tangent
spaces (Singer and Wu 2012). Quantum natural language processing
likewise provides precedents for representing semantic composition with
Hilbert spaces, category theory, tensor products, complex states, and
quantum operators (Coecke, Sadrzadeh, and Clark 2010; Coecke et
al.~2020).

Rather than merely juxtaposing these ideas, this paper integrates
\textbf{gauge-covariant phases of semantic states, a wavelength-based
hierarchical scale, a low-frequency discourse direction,
quantum-spectral selection, attention, and hallucination risk} into a
single operator framework.

\subsection{1.2 Terminology and Scope of
Claims}\label{terminology-and-scope-of-claims}

We call the overall model \textbf{QuantumPhaseNet} and its core
attention mechanism \textbf{WavePhase Attention}. The present work
extends our earlier WavePhaseNet framework, which constructed semantic
conceptual hierarchies by discrete Fourier analysis of transformer
representations (Kasubuchi and Fukiya 2026). Whereas that work operated
on scalar frequency bands in a fixed frame, the present paper replaces
the frame-dependent phase by a gauge-covariant one, introduces a
connection and curvature on a complex semantic bundle, and adds spectral
selection and a calibratable risk. The term \emph{quantum} refers to two
distinct computational realizations.

\begin{enumerate}
\def\labelenumi{\arabic{enumi}.}
\tightlist
\item
  \textbf{Quantum-executable realization:} state preparation, QFT, QPE,
  a reversible oracle, amplitude amplification, uncomputation, and
  measurement are implemented as a quantum circuit.
\item
  \textbf{Classical quantum-inspired realization:} the same spectral
  transforms, phase masks, reflections, and amplitude reweighting are
  implemented with classical linear algebra.
\end{enumerate}

The use of complex numbers, unitary matrices, or Born-type normalization
does not by itself produce a physical quantum speedup. We do not infer
exponential acceleration from the gate complexity of the QFT alone.
Total complexity must include state preparation, block encoding of the
Hamiltonian, oracle construction, error correction, and measurement
readout (Biamonte et al.~2017; Schuld and Killoran 2019; Gilyén et
al.~2019).

The results proved in this paper are conditional statements about the
internal consistency of the mathematical system as defined. The
following correspondences are empirical hypotheses and are therefore
falsifiable through experiment:

\begin{itemize}
\tightlist
\item
  longer wavelengths correspond to higher-level or more abstract
  concepts;
\item
  low-frequency components represent the correct discourse direction of
  a text;
\item
  connection curvature identifies harmful semantic deviation;
\item
  geometric risk predicts factual error and can be calibrated; and
\item
  the quantum-executable realization provides a practical advantage over
  classical approximations.
\end{itemize}

\subsection{1.3 Main Contributions}\label{main-contributions}

The paper makes the following ten contributions.

\begin{enumerate}
\def\labelenumi{\arabic{enumi}.}
\tightlist
\item
  It distinguishes phase angle, phase-path length, angular frequency,
  and wavelength, and defines semantic wavelength from a
  \textbf{covariant phase rate}.
\item
  It defines phase coherence not by the naive difference
  \(\varphi_i-\varphi_j\), but by a gauge-invariant quantity that
  includes parallel transport by a connection.
\item
  It converts the total preorder induced by a single wavelength into a
  total order on a quotient set, and constructs a conceptual partial
  order and distributive lattice from multiple wavelength axes.
\item
  It defines a discourse-direction vector using the Fréchet mean,
  logarithmic maps, graph low-pass filters, and the connection
  Laplacian.
\item
  It specifies the register structure of an external QFT, QPE, a
  target-projector oracle, amplitude amplification, uncomputation, and
  an external inverse QFT.
\item
  It defines WavePhase Attention by combining a causal mask, covariant
  phase, wavelength, direction, geodesic distance, and evidential
  support.
\item
  It establishes local gauge invariance, unitarity of the quantum block,
  the exact amplitude-amplification probability, and boundedness and
  conditional stability of the attention output.
\item
  It formulates hallucination as a nonnegative risk composed of
  geometric deviation, evidence deficiency, and semantic uncertainty,
  while separating calibration from abstention.
\item
  It unifies mathematical results, a minimally implementable model,
  falsifiable experiments, and a complexity analysis that does not
  assume quantum advantage.
\item
  It connects the theory to reproducible prototype experiments through
  the Validation Studio.
\end{enumerate}

The Validation Studio visualizes and compares, under a common
configuration, the processing pipeline of Section 14.1, the classical
complexity analysis of Section 15.1, and the five research questions of
Section 16.1. In this way, the theoretical hypotheses are connected to a
reproducible prototype experiment.

\section{2. Related Work}\label{related-work}

\subsection{2.1 Hierarchical Representation and Semantic
Geometry}\label{hierarchical-representation-and-semantic-geometry}

Symmetric distance alone is insufficient to represent hypernymy and
other directed inclusion relations. Order Embeddings use coordinate-wise
order constraints to encode asymmetric inclusion (Vendrov et al.~2016),
whereas Poincaré Embeddings match the volume growth of tree hierarchies
to that of negatively curved spaces (Nickel and Kiela 2017). The
wavelength order proposed here is not intended to replace these methods;
it is an additional hierarchical coordinate derived from spectral scale.
Concept inclusion should therefore not be inferred from wavelength
alone, but combined, when appropriate, with multi-axis orders,
hyperbolic geometry, and supervised inclusion losses.

\subsection{2.2 Graph Spectra and the Connection
Laplacian}\label{graph-spectra-and-the-connection-laplacian}

On a semantic graph constructed from a token sequence, low-eigenvalue
Laplacian modes represent components that vary smoothly over the graph.
In addition to scalar graph Fourier analysis, comparisons between
vectors in different tangent spaces require parallel transport, a
problem addressed by the connection Laplacian (Singer and Wu 2012). Our
discourse direction is therefore constructed not only from scalar
weights but also from low-frequency components of vector signals
transported to a common reference tangent space.

\subsection{2.3 Quantum Spectral
Processing}\label{quantum-spectral-processing}

QFT, QPE, and Grover search respectively provide a basis transform,
estimation of eigenphases of a unitary operator, and amplification of a
marked subspace (Coppersmith 1994; Kitaev 1995; Grover 1996; Nielsen and
Chuang 2010). More generally, quantum singular-value transformation
applies polynomial spectral filters to block-encoded operators (Gilyén
et al.~2019). We first use discrete band selection by QPE and amplitude
amplification as the central construction, and position QSVT as a
smoother alternative.

\subsection{2.4 Hallucination, Evidence, and
Calibration}\label{hallucination-evidence-and-calibration}

Retrieval-augmented generation connects generation to an external
nonparametric memory (Lewis et al.~2020). TruthfulQA, FActScore,
LongFact/SAFE, and FEVER provide benchmarks for truthfulness, evidential
support, and long-form factuality (Lin, Hilton, and Evans 2022; Min et
al.~2023; Wei et al.~2024; Thorne et al.~2018). SelfCheckGPT exploits
inconsistency across multiple generations (Manakul, Liusie, and Gales
2023), and semantic entropy measures uncertainty over semantic
equivalence classes rather than surface strings (Farquhar et al.~2024).
The geometric risk proposed here does not replace these methods; it
explicitly incorporates evidence consistency and semantic uncertainty as
separate terms. A probabilistic interpretation requires calibration, and
abstention guarantees may be combined with methods such as conformal
risk control (Angelopoulos et al.~2022; Abbasi-Yadkori et al.~2024).

\section{3. Mathematical Setting}\label{mathematical-setting}

\subsection{3.1 Sentences, Embeddings, and a Semantic
Manifold}\label{sentences-embeddings-and-a-semantic-manifold}

Let \(\mathcal{V}\) be a vocabulary and let a sentence be

\[S = \left( w_{1},\ldots,w_{n} \right) \in \mathcal{V}^{n}\]

Define the token embedding and positional representation by

\[x_{i} = e\left( w_{i} \right) + p_i^{\mathrm{pos}} \in \mathbb{R}^{d},\quad\quad X_{S} = \left\lbrack x_{1}^{\top};\ldots;x_{n}^{\top} \right\rbrack \in \mathbb{R}^{n \times d}\]

respectively.

\textbf{Assumption 1 (local semantic manifold).} Let
\(U \subset \mathbb{R}^{d}\) be open. Assume that there exist an
\(m\)-dimensional smooth manifold \(\mathcal{M}\) and a smooth map
\(\Phi:U\rightarrow\mathcal{M}\) such that
\(\operatorname{rank}D\Phi=m\) in a neighborhood of the observed data.
The semantic point associated with token \(i\) is \(z_i=\Phi(x_i)\).

This assumption does not require the entire embedding space to form one
global smooth semantic manifold. In implementation, sentence- or
segment-level local charts are treated as an atlas and patched together.

Define the local sentence region by

\[\mathcal{M}_{S} = \overline{\underset{i = 1}{\bigcup^{n}}B_{g}\left( z_{i},r_{i} \right)}\]

or by the geodesic convex hull
\(\operatorname{gconv}\{z_1,\ldots,z_n\}\) of the point cloud.

\subsection{3.2 Learned Metric}\label{learned-metric}

The semantic metric \(g\) may be a fixed Euclidean metric. A learnable
positive-definite metric can instead be defined by

\[G_{\theta}(x) = J_{f_{\theta}}(x)^{\top}J_{f_{\theta}}(x) + \varepsilon_{g}I_{d},\quad\quad\varepsilon_{g} > 0\]

which guarantees \(v^{\top}G_{\theta}(x)v>0\) for every nonzero \(v\).
For a curve \(\gamma:[0,1]\rightarrow\mathcal{M}\), its length and the
induced geodesic distance are

\[L_{g}(\gamma) = \int_{0}^{1}\sqrt{g_{\gamma(t)}\left( \dot{\gamma}(t),\dot{\gamma}(t) \right)}\, dt,\]

\[d_{g}(p,q) = \inf_{\gamma(0) = p,\,\gamma(1) = q}L_{g}(\gamma)\]

A geodesic satisfies \(\nabla^{g}_{\dot\gamma}\dot\gamma=0\) for the
Levi-Civita connection \(\nabla^{g}\).

\subsection{3.3 Complex Semantic Bundle}\label{complex-semantic-bundle}

Let the internal semantic degrees of freedom be represented by a
rank-\(r\) Hermitian vector bundle

\[\pi:E\rightarrow\mathcal{M},\quad\quad E_{p} \simeq \mathbb{C}^{r}\]

equipped with a unitary connection \(\nabla\). A semantic state is a
section

\[\Psi_{t}\in\mathcal{H}=L^{2}\left( \mathcal{M},E \right),\quad\quad \parallel \Psi_{t} \parallel_{\mathcal{H}}^{2} = \int_{\mathcal{M}}^{} \parallel \Psi_{t}(p) \parallel^{2}\, d\mu(p) = 1\]

and the semantic mass assigned by the model to a region
\(B\subset\mathcal{M}\) is

\[\Pr_{\Psi_{t}}(B) = \int_{B}^{} \parallel \Psi_{t}(p) \parallel^{2}\, d\mu(p)\]

This is normalized mass internal to the model; it is not an empirical
probability that an answer is correct.

\section{4. Gauge-Covariant Phase, Phase Accumulation, and Semantic
Wavelength}\label{gauge-covariant-phase-phase-accumulation-and-semantic-wavelength}

\subsection{4.1 Phase Angle and Covariant Phase
Rate}\label{phase-angle-and-covariant-phase-rate}

Locally, a nonzero semantic state may be written as

\[\psi(t) = A(t)e^{i\varphi(t)}u(t),\quad\quad A(t) > 0,\quad\quad \parallel u(t) \parallel = 1\]

but \(\varphi\) depends on the local frame. Consequently, the naive
derivative \(\dot\varphi\) is not invariant under a local gauge
transformation.

Consider a curve \(\gamma(t)\in\mathcal{M}\) following generation or
reading, and the normalized state
\(\widehat\psi(t)=\psi(t)/\|\psi(t)\|\). Define the covariant derivative
along the curve by

\[D_{t}\widehat{\psi} = \nabla_{\dot{\gamma}(t)}\widehat{\psi}\]

\textbf{Definition 2 (covariant phase rate).} The covariant phase rate
along \(\gamma\) is

\[\Omega_\gamma(t):=\operatorname{Im}\langle\widehat\psi(t),D_t\widehat\psi(t)\rangle\]

In a local \(U(1)\) representation with \(D_t=d/dt+iA_t\), if
\(\widehat\psi=e^{i\varphi}u\) and \(u\) is parallel, then
\(\Omega_\gamma=\dot\varphi+A_t\). Thus, the magnitude of phase
generated by an angle and elapsed time in the motivating theory is made
precise by the following accumulated quantity.

\textbf{Definition 3 (accumulated covariant phase).} The phase-path
length over \([0,T]\) is

\[\ell_{\varphi,\gamma}[0,T]=\int_0^T |\Omega_\gamma(t)|\,dt\]

The signed phase accumulation is
\(\Phi_\gamma[0,T]=\int_0^T\Omega_\gamma(t)\,dt\).

\subsection{4.2 Semantic Wavelength}\label{semantic-wavelength}

\textbf{Definition 4 (semantic wavelength).} For a positive semantic
propagation speed \(c_s(\gamma(t),t)>0\) and a stabilization constant
\(\varepsilon_\omega>0\), define

\[\boxed{\lambda_{\gamma}(t) = \frac{2\pi c_{s}\left( \gamma(t),t \right)}{\left| \Omega_{\gamma}(t) \right| + \varepsilon_{\omega}}}\quad\quad\text{(Wavelength)}\]

as the semantic wavelength. In normalized units \(c_s\equiv1\), slow
phase variation produces a long wavelength and rapid phase variation
produces a short wavelength.

The phase angle \(\varphi\), accumulated phase \(\ell_\varphi\), and
wavelength \(\lambda\) are distinct objects. The phase angle is a
coordinate on the circle, accumulated phase is the total variation along
time, and wavelength is the inverse scale of the covariant phase rate.
Because the motivating use of \emph{phase} may refer to any of these, we
distinguish them explicitly throughout the paper.

\subsection{4.3 Gauge Invariance}\label{gauge-invariance}

Consider the local \(U(1)\) gauge transformation

\[\widehat{\psi}'(t) = e^{i\chi(t)}\widehat{\psi}(t),\quad\quad A_{t}' = A_{t} - \dot{\chi}(t)\]

\textbf{Theorem 5 (local gauge invariance of covariant phase rate and
wavelength).} Under the above transformation,

\[\Omega_{\gamma}'(t) = \Omega_{\gamma}(t),\quad\quad\lambda_{\gamma}'(t) = \lambda_{\gamma}(t)\]

\textbf{Proof.} The connection transformation law gives
\(D_t'\widehat\psi'=e^{i\chi}D_t\widehat\psi\). By unitary invariance of
the Hermitian inner product,

\[\langle\widehat{\psi}',D_{t}'\widehat{\psi}'\rangle = \langle\widehat{\psi},D_{t}\widehat{\psi}\rangle.\]

Taking imaginary parts yields \(\Omega_\gamma'=\Omega_\gamma\), and the
definition then gives \(\lambda_\gamma'=\lambda_\gamma\). \(\square\)

\subsection{4.4 Covariant Phase Coherence Between Two
Points}\label{covariant-phase-coherence-between-two-points}

The fibers \(E_{z_i}\) and \(E_{z_j}\) over different points are not the
same vector space. Let \(\Gamma_{j\to i}\) be a selected path and let
\(P^{\nabla}_{j\to i}:E_{z_j}\to E_{z_i}\) denote parallel transport
along that path.

\textbf{Definition 6 (covariant phase coherence).} For normalized states
\(\widehat\psi_i\) and \(\widehat\psi_j\), define

\[C_{ij}^{\nabla}=\operatorname{Re}\langle\widehat\psi_i,P_{j\to i}^{\nabla}\widehat\psi_j\rangle\in[-1,1]\]

and define wavelength compatibility by

\[C_{ij}^{\lambda}=\exp\left[-\frac{(\log\lambda_i-\log\lambda_j)^2}{2\sigma_\lambda^2}\right]\in(0,1]\]

\textbf{Proposition 7 (gauge invariance of covariant coherence).} If the
fiber frames are transformed by \(g_i,g_j\in U(r)\) and parallel
transport is transformed as \(P'_{j\to i}=g_iP_{j\to i}g_j^{-1}\), then
\(C_{ij}^{\nabla}\) is invariant.

This definition removes the implicit requirement of a global common
frame imposed by the naive phase difference
\(\cos(\varphi_i-\varphi_j)\). When path dependence matters, one may fix
a shortest geodesic, the path induced by textual order, or a learned
reference path.

\section{5. Semantic Concept
Hierarchy}\label{semantic-concept-hierarchy}

\subsection{5.1 Total Preorder Induced by a Single
Wavelength}\label{total-preorder-induced-by-a-single-wavelength}

For a positive scale field \(\lambda:\mathcal{M}\to\mathbb{R}_{>0}\),
define

\[p \precsim_{\lambda}q\quad \Leftrightarrow \quad\lambda(p) \leq \lambda(q)\]

This relation is reflexive, transitive, and total, but it is not
antisymmetric because distinct points may share the same wavelength. It
is therefore a total preorder.

\textbf{Definition 8 (wavelength equivalence classes).} Let
\(p\sim_\lambda q\) if and only if \(\lambda(p)=\lambda(q)\), and set
\(\mathcal{C}_\lambda=\mathcal{M}/\sim_\lambda\). With a tolerance,
quantize by

\[b_{\varepsilon}(p) = \left\lfloor \frac{\log\lambda(p)}{\varepsilon} \right\rfloor,\quad\quad p \sim_{\varepsilon}q \Leftrightarrow b_{\varepsilon}(p) = b_{\varepsilon}(q)\]

so that transitivity is preserved.

\textbf{Definition 9 (hierarchical order on the quotient).}

\[\lbrack p\rbrack \preccurlyeq_{\lambda}\lbrack q\rbrack\quad \Leftrightarrow \quad\lambda(p) \leq \lambda(q)\]

\textbf{Theorem 10 (total order and distributive lattice on the
quotient).} The pair \((\mathcal{C}_\lambda,\preccurlyeq_\lambda)\) is a
totally ordered set, and for every pair of elements,

\[[p]\wedge[q]=\arg\min_{[r]\in\{[p],[q]\}}\lambda(r),\qquad [p]\vee[q]=\arg\max_{[r]\in\{[p],[q]\}}\lambda(r)\]

exist. Hence the quotient forms a distributive lattice under the binary
meet and join operations.

\textbf{Remark.} This does not mean that every subset has a supremum and
infimum, as would be required for a complete lattice. Nor is agreement
between wavelength order and lexical entailment a theorem; that
correspondence remains empirical.

\subsection{5.2 Multi-Axis Wavelengths and a Conceptual Partial
Order}\label{multi-axis-wavelengths-and-a-conceptual-partial-order}

Polysemy, attribute hierarchies, meronymy, and causal abstraction cannot
generally be represented by a single real number. Introduce \(k\)
gauge-invariant scales

\[\boldsymbol\lambda(p) = \left( \lambda^{(1)}(p),\ldots,\lambda^{(k)}(p) \right) \in \mathbb{R}_{> 0}^{k}\]

and define on the quotient

\[\lbrack p\rbrack \preccurlyeq_{\boldsymbol\lambda}\lbrack q\rbrack\quad \Leftrightarrow \quad\lambda^{(a)}(p) \leq \lambda^{(a)}(q)\quad(a = 1,\ldots,k)\]

\textbf{Proposition 11 (sufficient condition for a multi-axis
distributive lattice).} If the image \(\boldsymbol\lambda(\mathcal{M})\)
is closed under componentwise \(\min\) and \(\max\), then the quotient
becomes a distributive lattice with

\[\boldsymbol\lambda([p]\wedge[q])=\min\bigl(\boldsymbol\lambda([p]),\boldsymbol\lambda([q])\bigr),\]

\[\boldsymbol\lambda([p]\vee[q])=\max\bigl(\boldsymbol\lambda([p]),\boldsymbol\lambda([q])\bigr)\]

In implementation, for a supervised inclusion pair \((u,v)\), we fix the
sign convention that the scale of the subordinate concept \(u\) does not
exceed the scale of the superordinate concept \(v\) in any coordinate,
and combine this convention with an Order-Embedding-type asymmetric
loss. The orientation of larger and smaller wavelengths is not
arbitrary; it must be defined consistently for each dataset.

\section{6. Discourse Direction and Spectral
Hierarchy}\label{discourse-direction-and-spectral-hierarchy}

\subsection{6.1 Semantic Graph}\label{semantic-graph}

Construct a weighted graph \(G_S=(V,E,W)\) from the semantic points
\(z_i\) in a sentence. For example,

\[W_{ij} = \mathbf{1}\left\lbrack (i,j) \in E \right\rbrack\exp\left( - \frac{d_{g}\left( z_{i},z_{j} \right)^{2}}{2\sigma_{g}^{2}} \right)\exp\left( - \frac{|i - j|^{2}}{2\sigma_{p}^{2}} \right)\]

The edge set \(E\) may combine causal neighborhoods, \(k\)-nearest
neighbors, syntactic dependency edges, coreference edges, and
retrieval-evidence edges. Let \(D\) be the degree matrix and define the
normalized graph Laplacian

\[L=I-D^{-1/2}WD^{-1/2}=U\operatorname{diag}(\nu_0,\ldots,\nu_{n-1})U^\top\]

Its eigenvalues satisfy \(0=\nu_0\leq\nu_1\leq\cdots\), and low
eigenvalues represent globally smooth components on the graph.

\subsection{6.2 Reference Point and Vector
Signal}\label{reference-point-and-vector-signal}

Define the sentence reference point as the Fréchet mean

\[z_*=\arg\min_{z\in\mathcal M_S}\sum_{i=1}^n\rho_i d_g(z,z_i)^2,\qquad \rho_i\ge0,\quad\sum_i\rho_i=1\]

and obtain the tangent-space vectors through the logarithmic map

\[v_{i} = \log_{z_{*}}\left( z_{i} \right) \in T_{z_{*}}\mathcal{M}\]

If the logarithmic map is not unique, for example at a cut locus,
restrict the computation to a locally convex region or to a learned
chart.

For the vector signal \(V=[v_1;\ldots;v_n]\), define the low-pass filter
\(h_{\mathrm{low}}(L)\) by

\[V_{\mathrm{low}}=h_{\mathrm{low}}(L)V,\qquad h_{\mathrm{low}}(\nu)=\exp(-\tau\nu)\]

or by projection onto the first \(K\) low-frequency modes.

\subsection{6.3 Discourse-Direction
Vector}\label{discourse-direction-vector}

\textbf{Definition 12 (discourse-direction vector).} For sentence-level
importance weights \(a_i\geq0\), define

\[{\widetilde{\Theta}}_{S} = \sum_{i = 1}^{n}a_{i}\left( V_{low} \right)_{i},\quad\quad\Theta_{S} = \frac{{\widetilde{\Theta}}_{S}}{\parallel {\widetilde{\Theta}}_{S} \parallel_{g} + \varepsilon_{\Theta}} \in T_{z_{*}}\mathcal{M}\]

The term \emph{direction angle} generally refers here to a unit vector
in a tangent space rather than to a single scalar angle.

To compare this direction with a candidate direction
\(u_t\in T_{z_t}\mathcal{M}\) at a generation point \(z_t\),
parallel-transport \(\Theta_S\) to \(z_t\) and define

\[A_{t}^{\mathrm{dir}} = \frac{g_{z_{t}}\left( u_{t},P_{* \rightarrow t}\Theta_{S} \right)}{\left( \parallel u_{t} \parallel_{g} + \varepsilon \right)\left( \parallel P_{* \rightarrow t}\Theta_{S} \parallel_{g} + \varepsilon \right)} \in \lbrack - 1,1\rbrack\]

as the directional alignment score.

\textbf{Remark.} Low frequency does not necessarily represent the
correct intention. Repeated misinformation, long quotations, and topic
fixation may also become low-frequency components. The intended
direction must therefore be identified using summary supervision,
instruction representations, evidence graphs, and contrastive losses.

\subsection{6.4 Generalization with the Connection
Laplacian}\label{generalization-with-the-connection-laplacian}

When each \(v_i\) is retained in its own tangent space
\(T_{z_i}\mathcal{M}\), use parallel transport \(P_{j\to i}\) along edge
\((i,j)\) and define

\[\left( \Delta^{\nabla}v \right)_{i} = \sum_jW_{ij}\left( v_{i} - P_{j \rightarrow i}v_{j} \right)\]

This is a discrete connection Laplacian and reduces the distortion
introduced by transporting every vector to one reference point. A
minimal implementation may use an ordinary graph Laplacian and one
reference tangent space, whereas an extended implementation can use the
connection Laplacian directly.

\section{7. Fiber Bundles, Connections, and
Curvature}\label{fiber-bundles-connections-and-curvature}

\subsection{7.1 Local Gauges}\label{local-gauges}

Over a local coordinate patch \(U_\alpha\), let
\(E|_{U_\alpha}\simeq U_\alpha\times\mathbb{C}^r\), with transition
functions \(g_{\alpha\beta}:U_\alpha\cap U_\beta\to U(r)\). For a local
connection one-form \(A_\alpha\),

\[\nabla = d + A_{\alpha},\quad\quad A_{\beta} = g_{\alpha\beta}A_{\alpha}g_{\alpha\beta}^{- 1} - \left( dg_{\alpha\beta} \right)g_{\alpha\beta}^{- 1}\]

Parallel transport along a curve \(\gamma\) is

\[P_\gamma=\mathcal P\exp\left(-\int_\gamma A\right)\]

\subsection{7.2 Curvature and Holonomy}\label{curvature-and-holonomy}

The curvature two-form is

\[F_{\nabla} = dA + A \land A\]

and the holonomy around a small closed curve \(\Gamma\) is

\[\operatorname{Hol}(\Gamma)=\mathcal P\exp\left(-\oint_\Gamma A\right)\approx I-\int_\Sigma F_\nabla\]

Semantically, this quantity measures how far an internal semantic state
fails to return to itself after traversing a closed contextual loop.

\textbf{Definition 13 (curvature risk).} For a neighborhood \(B_t\) of a
generation point \(z_t\), define

\[R_{t}^{\mathrm{curv}} = \frac{1}{\mu\left( B_{t} \right)}\int_{B_t} \parallel F_{\nabla}(z) \parallel_{F}^{2}\, d\mu(z)\]

Large curvature alone does not imply error. Polysemy, metaphor,
viewpoint changes, and legitimate topic transitions may also increase
curvature. Curvature must therefore be interpreted jointly with
direction, evidence, and uncertainty.

\section{8. Semantic Hamiltonian}\label{semantic-hamiltonian}

\subsection{8.1 Finite Semantic
Register}\label{finite-semantic-register}

An implementation uses \(N=2^m\) basis states. If the sentence length is
\(n<N\), masked padding is applied. Let the complex feature matrix be

\[Z_{j\ell}=(W_{\mathrm{Re}}x_j)_\ell+i(W_{\mathrm{Im}}x_j)_\ell\in\mathbb C,\qquad Z\in\mathbb C^{N\times c}\]

and consider amplitude encoding into a data register and a channel
register:

\[|\psi_X\rangle=\frac{1}{\lVert Z\rVert_F}\sum_{j=0}^{N-1}\sum_{\ell=0}^{c-1}Z_{j\ell}|j\rangle|\ell\rangle\]

If the channel dimension \(c\) is not a power of two, the channel
register is padded as well. The external transform \(F_N\) acts only on
the first register. Because preparing a general amplitude-encoded state
may dominate an actual quantum circuit, basis encoding, angle encoding,
and low-rank state preparation must also be compared.

\subsection{8.2 Self-Adjoint Operator}\label{self-adjoint-operator}

Define a finite-dimensional semantic Hamiltonian by

\[H_{\mathrm{sem}} = \alpha_{L}L_{\nabla} + V_{\mathrm{task}} + \alpha_{h}R_{\mathrm{hier}} + \alpha_{e}R_{\mathrm{evid}} + \alpha_{c}R_{\mathrm{curv}}\]

Each component is constructed as a Hermitian matrix. A learned
non-Hermitian matrix \(B\) is symmetrized as \((B+B^\dagger)/2\).

\textbf{Assumption 14 (bounded self-adjointness).} Assume
\(H_{\mathrm{sem}}=H_{\mathrm{sem}}^\dagger\) and
\(\|H_{\mathrm{sem}}\|_2\leq\Lambda_H\).

By the spectral theorem,

\[H_{\mathrm{sem}} = \sum_{a = 0}^{N - 1}E_{a}\left| E_{a}\rangle\langle E_{a} \right|.\]

To avoid phase aliasing in QPE, define the unitary

\[U_H=\exp\left(2\pi i\frac{H_{\mathrm{sem}}-E_{\min}I}{\Delta_E}\right),\qquad \Delta_E>E_{\max}-E_{\min}\]

whose eigenphases are

\[U_H|E_a\rangle=e^{2\pi i\phi_a}|E_a\rangle,\qquad \phi_a=\frac{E_a-E_{\min}}{\Delta_E}\in[0,1).\]

Without this explicit scaling, distinct eigenenergies may wrap to the
same phase.

\section{9. QFT--QPE--Oracle--Amplitude Amplification--Inverse
QFT}\label{qftqpeoracleamplitude-amplificationinverse-qft}

\subsection{9.1 Distinguishing Two Fourier
Transforms}\label{distinguishing-two-fourier-transforms}

Define the external Fourier transform on the \(N\) token coordinates by

\[F_N|j\rangle=\frac{1}{\sqrt N}\sum_{k=0}^{N-1}e^{2\pi ijk/N}|k\rangle\]

This transform maps token-position coordinates to frequency coordinates.

Standard QPE, by contrast, uses Hadamard gates, controlled powers
\(U_H^{2^r}\), and an \textbf{inverse QFT} on an auxiliary phase
register. Thus, the external \(F_N\) and the internal
\(F_{2^m}^\dagger\) of QPE act on different registers and serve
different roles. The processing sequence is

\[
\begin{aligned}
|\psi_X\rangle
&\xrightarrow{F_N}|\psi_\omega\rangle
\xrightarrow{\mathrm{QPE}(U_H)}\sum_a c_a|E_a\rangle|\widetilde\phi_a\rangle\\
&\xrightarrow{O_G,\,\mathrm{AA}}\text{amplified marked phase sector}
\xrightarrow{\mathrm{QPE}^{\dagger}}|\psi_\omega'\rangle
\xrightarrow{F_N^{\dagger}}|\psi_X'\rangle .
\end{aligned}
\]

\subsection{9.2 Precision of QPE}\label{precision-of-qpe}

With \(m\) auxiliary qubits, QPE records an \(m\)-bit approximation
\(\widetilde\phi_a\) of an eigenphase \(\phi_a\). Finite precision
causes leakage near phase-window boundaries. The oracle should therefore
use a margin \(\delta_\phi\) or a smooth window function rather than
only a hard threshold.

\subsection{9.3 Target Subspace and
Oracle}\label{target-subspace-and-oracle}

Construct a reversible predicate from the target direction, wavelength
band, evidence, and prohibited conditions:

\[g(a,e,d,h) \in \{ 0,1\}\]

Here \(a\) is an eigenphase label, \(e\) an evidence flag, \(d\) a
direction-alignment bin, and \(h\) a hierarchy bin. Define the target
projector by

\[\Pi_G=\sum_{g(a,e,d,h)=1}|a,e,d,h\rangle\langle a,e,d,h|\]

and the reflection oracle by

\[O_{G} = I - 2\Pi_{G}\]

\textbf{Theorem 15 (self-adjointness and involution of the oracle).}
\(O_G^\dagger=O_G\) and \(O_G^2=I\).

\textbf{Proof.} Since \(\Pi_G\) is an orthogonal projector,
\(\Pi_G^\dagger=\Pi_G\) and \(\Pi_G^2=\Pi_G\). Hence

\[O_{G}^{\dagger} = I - 2\Pi_{G} = O_{G},\quad\quad O_{G}^{2} = I - 4\Pi_{G} + 4\Pi_{G}^{2} = I.\]

\subsection{9.4 Amplitude Amplification from a General Initial
State}\label{amplitude-amplification-from-a-general-initial-state}

Let \(|\psi_0\rangle\) be the normalized state after QPE, and let its
initial marked mass be

\[a=\lVert\Pi_G|\psi_0\rangle\rVert^2\in(0,1),\qquad \theta=\arcsin\sqrt a\]

Define reflection about the initial state by

\[S_{\psi} = 2\left| \psi_{0}\rangle\langle\psi_{0} \right| - I\]

and one amplitude-amplification iteration by \(Q=S_\psi O_G\).

\textbf{Theorem 16 (exact success probability of amplitude
amplification).} After \(k\) iterations, the probability of measuring
the target subspace is

\[P_k=\lVert\Pi_GQ^k|\psi_0\rangle\rVert^2=\sin^2((2k+1)\theta).\]

Moreover, if

\[k_*=\operatorname{round}\left(\frac{\pi}{4\theta}-\frac12\right)\]

then \(P_{k_*}\geq1-a\).

\textbf{Proof sketch.} The state remains in the two-dimensional plane
spanned by its marked and unmarked components, and \(Q\) acts in that
plane as a rotation by \(2\theta\). The nearest-integer choice makes the
final angle differ from \(\pi/2\) by at most \(\theta\), so
\(P_{k_*}\geq\cos^2\theta=1-a\). \(\square\)

If the marked mass \(a\) is unknown, a fixed number of iterations may
overshoot. Implementations should compare iteration-count estimation,
randomized iteration, fixed-point amplitude amplification, and smooth
spectral filters.

\subsection{9.5 Unitarity of the Full
Block}\label{unitarity-of-the-full-block}

Conceptually define the quantum selection block by

\[\mathcal U_{\mathrm{QPN}}=F_N^\dagger\,\mathrm{QPE}^\dagger\,Q^k\,\mathrm{QPE}\,F_N\]

\textbf{Theorem 17 (unitarity of the QuantumPhaseNet quantum block).} If
\(F_N\), QPE, \(O_G\), and \(S_\psi\) are unitary, then
\(\mathcal{U}_{\mathrm{QPN}}\) is unitary.

\textbf{Proof.} Products and adjoints of unitary operators are unitary.
\(\square\)

This theorem does not imply a complexity advantage. If measurement or
post-selection is inserted in the middle of the procedure, the overall
process is generally nonunitary and should instead be described as a
quantum channel.

\subsection{9.6 Classical Quantum-Inspired
Approximation}\label{classical-quantum-inspired-approximation}

In the classical realization, approximate the target projector \(\Pi_G\)
by eigendecomposition or a Krylov method and apply

\[\widetilde z=\frac{(I+\eta\Pi_G)z}{\lVert(I+\eta\Pi_G)z\rVert_2},\qquad\eta\ge0\]

or use a Chebyshev polynomial \(p_K(H_{\mathrm{sem}})\). This does not
claim Grover's quadratic speedup, but it explicitly increases the weight
assigned to the target subspace.

\section{10. WavePhase Attention}\label{wavephase-attention}

\subsection{10.1 Geometry- and Phase-Corrected
Logits}\label{geometry--and-phase-corrected-logits}

For every token pair \((i,j)\), add the following features to the
standard attention logit:

\begin{itemize}
\tightlist
\item
  covariant phase coherence \(C_{ij}^{\nabla}\);
\item
  wavelength compatibility \(C_{ij}^{\lambda}\);
\item
  discourse-direction alignment \(A_{ij}^{\mathrm{dir}}\);
\item
  geodesic proximity \(D_{ij}^{g}=d_g(z_i,z_j)^2\);
\item
  evidential support \(E_{ij}^{\mathrm{evid}}\in[0,1]\); and
\item
  a causal/padding mask \(M_{ij}\in\{0,-\infty\}\).
\end{itemize}

\textbf{Definition 18 (WavePhase logit).} Parameterizing nonnegative
coefficients with softplus, define

\[\boxed{\ell_{ij}^{\mathrm{WP}}=\frac{q_i^\top k_j}{\sqrt{d_k}}+\alpha_\varphi C_{ij}^{\nabla}+\alpha_\lambda C_{ij}^{\lambda}+\alpha_d A_{ij}^{\mathrm{dir}}-\alpha_g D_{ij}^{g}+\alpha_e E_{ij}^{\mathrm{evid}}+M_{ij}}\tag{WP}\]

The directional term \(A_{ij}^{\mathrm{dir}}\) may be asymmetric because
it measures whether key \(j\) lies along the discourse direction as
viewed from query \(i\). For example,

\[A_{ij}^{\mathrm{dir}} = \frac{g_{z_{i}}\left( \log_{z_{i}}z_{j},P_{* \rightarrow i}\Theta_{S} \right)}{\left( \parallel \log_{z_{i}}z_{j} \parallel_{g} + \varepsilon \right)\left( \parallel P_{* \rightarrow i}\Theta_{S} \parallel_{g} + \varepsilon \right)}.\]

\subsection{10.2 Attention Output}\label{attention-output}

\textbf{Definition 19 (WavePhase Attention).} Apply a row-wise softmax:

\[A_{ij}^{\mathrm{WP}}=\frac{\exp(\ell_{ij}^{\mathrm{WP}})}{\sum_{j':M_{ij'}=0}\exp(\ell_{ij'}^{\mathrm{WP}})},\qquad Y=A^{\mathrm{WP}}V\]

Integrate the quantum-selected feature \(Z_{\mathrm{QPN}}\) through a
residual connection:

\[Y_{\mathrm{out}}=Y+\beta_qW_qZ_{\mathrm{QPN}}\]

The initial value of \(\beta_q\) should be small to avoid instability
early in training.

\subsection{10.3 Complex-Valued Variant}\label{complex-valued-variant}

For complex-valued vectors \(V_j^{\mathbb{C}}\), define

\[{\widetilde{V}}_{j} = e^{i\vartheta_{j}}V_{j}^{\mathbb{C}},\quad\quad Y_{i}^{\mathbb{C}} = \sum_jA_{ij}^{\mathrm{WP}}{\widetilde{V}}_{j}\]

and map back to a real-valued model with

\[\rho(Y^{\mathbb C})=[\operatorname{Re}Y^{\mathbb C};\operatorname{Im}Y^{\mathbb C}]\]

or with a complex linear layer. This makes constructive and destructive
interference from phase differences explicit, at the cost of more
complex-valued parameters and a more difficult optimization problem.

\subsection{10.4 Invariance, Boundedness, and
Stability}\label{invariance-boundedness-and-stability}

\textbf{Theorem 20 (local gauge invariance of the WavePhase logit).} If
\(C_{ij}^{\nabla}\), \(\lambda_i\), \(\lambda_j\), geodesic distance,
directional inner products, and evidence scores are constructed from
gauge-invariant quantities, then \(\ell_{ij}^{\mathrm{WP}}\) and
\(A_{ij}^{\mathrm{WP}}\) are invariant under local unitary gauge
transformations.

\textbf{Theorem 21 (row-norm boundedness of the output).} If every value
vector satisfies \(\|v_j\|_2\leq B_V\), then every output row satisfies

\[\lVert Y_i\rVert_2\le B_V\]

\textbf{Proof.} Because \(A_{ij}^{\mathrm{WP}}\geq0\) and
\(\sum_jA_{ij}^{\mathrm{WP}}=1\), \(Y_i\) is a convex combination of the
value vectors. The result follows from the triangle inequality.
\(\square\)

\textbf{Assumption 22 (bounded regular region).} On the compact region
under consideration, assume that
\(Q,K,V,C^{\nabla},C^{\lambda},A^{\mathrm{dir}},D^g,E^{\mathrm{evid}}\)
are Lipschitz-continuous functions of the input \(X\) and that all these
quantities are bounded.

\textbf{Theorem 23 (local Lipschitz stability).} Under Assumption 22,
the map \(X\mapsto Y(X)\) is Lipschitz continuous on the compact region.

\textbf{Proof sketch.} The corrected logit is a finite sum of Lipschitz
maps. The Jacobian of softmax is \(J=\operatorname{diag}(p)-pp^\top\),
whose Euclidean operator norm is at most \(1/2\). Apply boundedness and
Lipschitz estimates to the product difference

\[A(X)V(X) - A(X')V(X') = \left\lbrack A(X) - A(X') \right\rbrack V(X) + A(X')\left\lbrack V(X) - V(X') \right\rbrack\]

. \(\square\)

The theorem does not guarantee adversarial robustness. Its assumptions
may fail if feature estimators are discontinuous, geodesics branch, or a
hard oracle boundary is differentiated through directly.

\section{11. Generation Dynamics}\label{generation-dynamics}

\subsection{11.1 Discrete Semantic
Trajectory}\label{discrete-semantic-trajectory}

Let \(z_t\in\mathcal{M}\) be the semantic point associated with
generation step \(t\), and define the local update

\[z_{t+1}=\operatorname{Exp}_{z_t}\!\left(\eta_tF_\theta(z_t,S,R_t)\right)\]

where \(R_t\) denotes retrieved evidence and \(F_\theta\) is a tangent
vector produced from WavePhase Attention, quantum-selected features, and
the language-model head.

\subsection{11.2 Energy with a Target
Direction}\label{energy-with-a-target-direction}

For a target set \(\mathcal{G}\), an evidence set \(\mathcal{E}_t\), and
a reference discourse direction \(\Theta_S\), define

\[\mathcal{E}_{t}(z) = \beta_{G}d_g(z,\mathcal G)^{2} + \beta_{P}d_{g}\left( z,\Gamma_{S} \right)^{2} + \beta_{D}\left( 1 - A^{\mathrm{dir}}(z) \right) + \beta_{E}R^{\mathrm{evid}}\left( z;\mathcal{E}_{t} \right) + \beta_{C}R^{\mathrm{curv}}(z)\]

The reference path \(\Gamma_S\) is not merely a shortest geodesic; it is
constrained to satisfy the instruction, factual, safety, and formatting
requirements.

The generation update can be interpreted as the Riemannian gradient step

\[z_{t+1}=\operatorname{Exp}_{z_t}\!\left(-\eta\,\operatorname{grad}\mathcal E_t(z_t)\right)\]

\textbf{Theorem 24 (conditional linear convergence).} Suppose
\(\mathcal{M}\) is a Hadamard manifold, the time-invariant energy
\(\mathcal{E}\) is geodesically \(\mu\)-strongly convex and
\(L\)-smooth, and \(0<\eta\leq1/L\). If \(z^*\) is its unique minimizer,
then

\[\mathcal E(z_t)-\mathcal E(z^*)\le(1-\mu\eta)^t[\mathcal E(z_0)-\mathcal E(z^*)]\]

\textbf{Proof sketch.} On a Hadamard manifold, the exponential map is
globally defined. The descent lemma for geodesically convex smooth
functions applies, and strong convexity yields a
Polyak--Łojasiewicz-type inequality that can be iterated. \(\square\)

We do not claim that natural-language semantic space is a Hadamard
manifold or that a generation energy is strongly convex. The theorem is
conditional: it identifies assumptions under which the framework would
guarantee stable tracking of a target.

\subsection{11.3 Time-Varying Targets and Tracking
Error}\label{time-varying-targets-and-tracking-error}

In practice, both retrieved evidence and the generation objective vary
over time. For a reference trajectory \(z_t^*\), define

\[\lVert\log_{z_t}(z_t^*)\rVert_g\]

as the tracking error, and separately evaluate target drift
\(d_g(z_{t+1}^*,z_t^*)\), local contraction, and discretization error.
In a model that permits topic transitions, the direction term should be
a piecewise-updated \(\Theta_{S,t}\) rather than a fixed vector.

\section{12. Hallucination Risk}\label{hallucination-risk}

\subsection{12.1 Scope of the Risk Model}\label{scope-of-the-risk-model}

Hallucinations arise from multiple causes, including missing knowledge,
retrieval failure, proposition-composition failure, reasoning failure,
citation error, ambiguous instructions, and decoding variability. The
risk defined here is not a causal explanation of all such mechanisms. In
particular, semantic uncertainty alone may fail to detect a fluent but
consistently wrong answer.

Decompose a generated text into atomic claims
\(c_{t,1},\ldots,c_{t,m_t}\), and associate an evidence set
\(\mathcal{E}_t\) with each claim.

\subsection{12.2 Five Deviation Terms}\label{five-deviation-terms}

\textbf{Directional deviation:}

\[R_{t}^{\mathrm{dir}} = 1 - A_{t}^{\mathrm{dir}}.\]

\textbf{Path deviation:}

\[R_{t}^{path} = d_{g}\left( z_{t},\Gamma_{S} \right)^{2}.\]

\textbf{Curvature deviation:}

\[R_t^{\mathrm{curv}}=\frac{1}{\mu(B_t)}\int_{B_t}\lVert F_\nabla\rVert_F^2\,d\mu.\]

\textbf{Evidence deficiency:} for a verifier
\(s_{\mathrm{sup}}(c,\mathcal{E})\in[0,1]\), define

\[R_{t}^{\mathrm{evid}} = \frac{1}{m_{t}}\sum_{a = 1}^{m_{t}}\left\lbrack 1 - s_{\sup}\left( c_{t,a},\mathcal{E}_{t} \right) \right\rbrack.\]

\textbf{Semantic uncertainty:} use either leakage from the quantum
target subspace or entropy over semantic classes:

\[R_t^{\mathrm{unc}} = 1 - \langle\psi_{t}\left| \Pi_{G} \right|\psi_{t}\rangle\]

or

\[R_t^{\mathrm{unc}}=-\sum_{c\in\mathcal C_t}p(c\mid S,R_t)\log p(c\mid S,R_t)\]

\subsection{12.3 Integrated Risk and
Calibration}\label{integrated-risk-and-calibration}

\textbf{Definition 25 (stepwise hallucination risk).} For nonnegative
coefficients \(\beta_\bullet\), define

\[\boxed{\mathcal{R}_{t} = \beta_{d}R_{t}^{\mathrm{dir}} + \beta_{p}R_{t}^{path} + \beta_{c}R_{t}^{\mathrm{curv}} + \beta_{e}R_{t}^{\mathrm{evid}} + \beta_{u}R_t^{\mathrm{unc}}}\quad\quad\text{(Risk)}\]

and define the sequence-level risk as the discounted sum

\[\mathcal{R}_{1:T} = \sum_{t = 1}^{T}\gamma^{T - t}\mathcal{R}_{t},\quad\quad 0 < \gamma \leq 1\]

The quantity \(\mathcal{R}_t\) is a score by definition, not the true
probability of error. On validation data, learn a logistic calibration
map

\[{\widehat{p}}_{t} = \sigma\left( a\mathcal{R}_{t} + b \right)\]

or isotonic regression, and evaluate the result with Brier score, ECE,
AUROC, and risk--coverage curves. Training, calibration, and final
evaluation splits must be kept separate.

\subsection{12.4 Selective Generation}\label{selective-generation}

For a threshold \(\tau\), define

\[\text{answer if }{\widehat{p}}_{t} \leq \tau,\quad\quad\text{retrieve/revise/abstain if }{\widehat{p}}_{t} > \tau\]

In addition to a single fixed threshold, compare conformal abstention
procedures targeting a specified risk level. Any guarantee must state
assumptions such as exchangeability, calibration-set size, and
monotonicity of the loss.

\subsection{12.5 A Limited Zero-Risk Convergence
Proposition}\label{a-limited-zero-risk-convergence-proposition}

\textbf{Proposition 26 (convergence to the reference path under an error
bound).} Suppose that on a compact set \(K\),

\[\mathcal{R}(z) \geq c\, d_{g}\left( z,\Gamma_{S} \right)^{2},\quad\quad c > 0\]

If \(z_t\in K\) and \(\mathcal{R}(z_t)\to0\), then
\(d_g(z_t,\Gamma_S)\to0\).

This proposition does not identify which point on the reference path is
approached, preserve the temporal ordering of the trajectory, or
guarantee the truth of generated claims. The error-bound assumption
itself is strong.

\section{13. Learning Objectives}\label{learning-objectives}

Define the total loss by

\[\begin{matrix}
\mathcal{L} = & \mathcal L_{\mathrm{LM}} + \lambda_{\varphi}\mathcal{L}_{\varphi} + \lambda_{h}\mathcal L_{\mathrm{hier}} + \lambda_{d}\mathcal L_{\mathrm{dir}} + \lambda_{g}\mathcal L_{\mathrm{geo}} \\
 & + \lambda_{c}\mathcal L_{\mathrm{curv}} + \lambda_{e}\mathcal L_{\mathrm{evid}} + \lambda_{u}\mathcal L_{\mathrm{unc}} + \lambda_{q}\mathcal L_{\mathrm{unit}} + \lambda_{cal}\mathcal L_{\mathrm{cal}}
\end{matrix}\]

\subsection{13.1 Phase Loss}\label{phase-loss}

For a supervised or contrastive pair \((i,j)\), use covariant coherence
rather than a naive angular difference:

\[\mathcal{L}_{\varphi} = \sum_{(i,j) \in \mathcal{P}^{+}}^{}\left( 1 - C_{ij}^{\nabla} \right) + \sum_{(i,j) \in \mathcal{P}^{-}}^{}\max\left( 0,C_{ij}^{\nabla} - m_{\varphi} \right)\]

\subsection{13.2 Hierarchy Loss}\label{hierarchy-loss}

For a supervised pair consisting of a subordinate concept \(u\) and a
superordinate concept \(v\), define

\[\mathcal L_{\mathrm{hier}}=\sum_{(u,v)}\sum_{a=1}^k[\max\{0,\log\lambda^{(a)}(u)-\log\lambda^{(a)}(v)+m_h\}]^2\]

The sign convention must be fixed over the entire dataset.
Reverse-direction negatives, transitive closure, and synonym equivalence
classes should be evaluated separately.

\subsection{13.3 Direction, Geometry, and Curvature
Losses}\label{direction-geometry-and-curvature-losses}

For a target direction \(\Theta_S^*\) obtained from a summary or
instruction, define

\[\mathcal L_{\mathrm{dir}} = 1 - g\left( \Theta_{S},\Theta_{S}^{*} \right)\]

Local distance preservation can be enforced with

\[\mathcal L_{\mathrm{geo}}=\sum_{(i,j)}[d_g(z_i,z_j)-d_{\mathrm{teacher}}(i,j)]^2\]

Curvature regularization should not force all curvature to zero.
Excluding supervised regions of legitimate topic transition, use

\[\mathcal L_{\mathrm{curv}}=\sum_t\omega_t^{\mathrm{stable}}R_t^{\mathrm{curv}}\]

\subsection{13.4 Evidence and Calibration
Losses}\label{evidence-and-calibration-losses}

For an evidential-support label \(y_{t,a}\in\{0,1\}\) of an atomic
claim, define

\[\mathcal L_{\mathrm{evid}}=-\sum_{t,a}[y_{t,a}\log s_{t,a}+(1-y_{t,a})\log(1-s_{t,a})]\]

and use the Brier loss for calibration:

\[\mathcal L_{\mathrm{cal}} = \frac{1}{N}\sum_{i = 1}^{N}\left( {\widehat{p}}_{i} - y_{i} \right)^{2}\]

\subsection{13.5 Unitary and Self-Adjoint
Regularization}\label{unitary-and-self-adjoint-regularization}

For a learned operator \(U_\theta\), add

\[\mathcal L_{\mathrm{unit}}=\lVert U_\theta^\dagger U_\theta-I\rVert_F^2\]

Self-adjointness of the Hamiltonian can be enforced exactly by
parameterizing \(H_\theta=(B_\theta+B_\theta^\dagger)/2\).

\section{14. Algorithms}\label{algorithms}

\subsection{14.1 Classical Quantum-Inspired
Realization}\label{classical-quantum-inspired-realization}

\textbf{Algorithm 1: Classical Quantum-Inspired QuantumPhaseNet}

\begin{enumerate}
\def\labelenumi{\arabic{enumi}.}
\tightlist
\item
  Embed the input text to obtain \(X_S\).
\item
  Construct \(G_\theta(x)\), the semantic points \(z_i\), and the local
  semantic graph \(G_S\).
\item
  Estimate \(\Theta_S\) from the Fréchet mean \(z_*\), logarithmic maps
  \(v_i\), and the low-frequency component \(V_{\mathrm{low}}\).
\item
  Estimate complex states, covariant phase rates \(\Omega_i\), semantic
  wavelengths \(\lambda_i\), and connection-aware coherence
  \(C_{ij}^{\nabla}\).
\item
  Construct a self-adjoint \(H_{\mathrm{sem}}\) and extract target
  spectral components with Lanczos, Chebyshev, or low-rank methods.
\item
  Reweight amplitudes using the approximate target projector
  \(\widetilde\Pi_G\).
\item
  Compute causal WavePhase Attention using Eq. (WP).
\item
  Generate candidates with the language-model head, decompose them into
  claims, verify evidence, and use \(\mathcal{R}_t\) to trigger
  retrieval, regeneration, or abstention.
\item
  Train end to end or by stages with the total loss.
\end{enumerate}

\subsection{14.2 Quantum-Executable
Realization}\label{quantum-executable-realization}

\textbf{Algorithm 2: Quantum-Executable Spectral Selection}

\begin{enumerate}
\def\labelenumi{\arabic{enumi}.}
\tightlist
\item
  Pad to \(N=2^m\) and prepare normalized semantic features as a quantum
  state.
\item
  Apply the external \(F_N\) to the data register.
\item
  Run QPE for \(U_H\) and write eigenphases to an auxiliary register.
\item
  Reversibly compute direction, wavelength, evidence, and
  prohibited-condition predicates, and apply \(O_G=I-2\Pi_G\).
\item
  Perform amplitude amplification using a known or estimated marked
  mass.
\item
  Uncompute the auxiliary predicates and QPE to disentangle the
  ancillas.
\item
  Apply the external \(F_N^\dagger\).
\item
  Measure only the probability of the marked subspace, low-order
  moments, or local observables, and return them as residual features to
  the classical Transformer.
\end{enumerate}

Reading out every amplitude may require \(\Omega(N)\) measurements. The
quantum realization should therefore avoid full-vector reconstruction
and instead return a small number of observables or connect directly to
a subsequent quantum-processing stage.

\section{15. Complexity and Numerical
Stability}\label{complexity-and-numerical-stability}

\subsection{15.1 Classical Complexity}\label{classical-complexity}

Standard dense attention requires \(O(n^2d)\) time and \(O(n^2)\) memory
for the attention matrix. All-pairs WavePhase corrections have the same
asymptotic order with a larger constant factor. For a sparse graph with
\(|E|=O(kn)\), graph construction is approximately \(O(knd)\), \(K\) low
eigenmodes can be estimated by Lanczos iterations in approximately
\(O(TK|E|)\), and a Chebyshev filter costs \(O(K|E|d)\).

To avoid all-pairs geodesic computation, compare local tangent-space
distances, neighborhood-only computation, Nystrom approximation, and
landmark methods. Parallel transport can be evaluated only on edges,
with long-range transport approximated by products along paths.

\subsection{15.2 Limits of the Quantum Complexity
Claim}\label{limits-of-the-quantum-complexity-claim}

A QFT on \(m=\log_2N\) qubits can be implemented with roughly \(O(m^2)\)
elementary gates and further reduced with an approximate QFT. End-to-end
cost, however, depends on (i) state preparation of semantic features,
(ii) implementation of \(H_{\mathrm{sem}}\) or \(U_H\), (iii) the
controlled evolution required for QPE precision \(\epsilon_\phi\), (iv)
reversible evaluation of the oracle, (v) \(O(1/\sqrt a)\)
amplitude-amplification iterations, (vi) error correction, and (vii)
readout. We therefore make no claim that merely replacing a DFT with a
QFT yields a speedup.

\subsection{15.3 Numerical Stability}\label{numerical-stability}

\begin{itemize}
\tightlist
\item
  Include \(\varepsilon_\omega>0\) in the wavelength definition to
  prevent divergence near zero frequency.
\item
  Use a covariant inner-product loss or a \(1-\cos\) loss rather than
  squared angular differences.
\item
  Clip \(\log\lambda\) to limit extreme scale ratios.
\item
  Constrain metric eigenvalues to \([\varepsilon_g,M_g]\).
\item
  Compute the Fréchet mean and logarithmic maps within a locally convex
  radius.
\item
  Relax hard spectral masks to sigmoid or raised-cosine windows during
  training.
\item
  Estimate \(a\) or use a fixed-point method to avoid
  amplitude-amplification overshoot.
\item
  Normalize complex states at every layer and separate the gradient
  scales of phase and amplitude.
\end{itemize}

\section{16. Experimental Method and Evaluation
Design}\label{experimental-method-and-evaluation-design}

\subsection{16.1 Research Questions}\label{research-questions}

\textbf{RQ1 (hierarchy).} Do the learned \(\lambda\) or
\(\boldsymbol\lambda\) values agree with known hypernym--hyponym
relations?

\textbf{RQ2 (discourse).} Does \(\Theta_S\) align with summaries,
instructions, and discourse centers, and does it reduce topic drift in
long-form generation?

\textbf{RQ3 (phase).} Does covariant phase coherence have greater
predictive power than ordinary cosine similarity between embeddings?

\textbf{RQ4 (evidence and hallucination).} Does \(\mathcal{R}_t\)
identify and calibrate errors better than entropy, maximum softmax
probability, SelfCheckGPT, semantic entropy, and related baselines?

\textbf{RQ5 (quantum selection).} Is spectral selection more effective
than a classical filter under the same computational budget, and is it
robust to noise in quantum-circuit simulation?

\subsection{16.2 Datasets}\label{datasets}

Hierarchy evaluation should use WordNet hypernym relations, graded
lexical entailment from HyperLex, and, where appropriate,
image--language Order-Embedding benchmarks. Factuality should be
evaluated with TruthfulQA, FEVER, FActScore biographies, and
LongFact/SAFE. To evaluate long-form directional stability, a dedicated
dataset should be constructed in which models generate multiple sections
from an instruction and experts annotate the intended discourse of each
paragraph and permissible topic-transition points.

For Japanese evaluation, a separate dataset should include
language-specific ellipsis, honorifics, omitted subjects, polysemy, and
variation in kanji spelling rather than merely translating English data.
Multilingual experiments should isolate differences caused by the NLI or
evidence verifier itself.

\subsection{16.3 Baselines and Ablations}\label{baselines-and-ablations}

At minimum, compare the following systems:

\begin{enumerate}
\def\labelenumi{\arabic{enumi}.}
\tightlist
\item
  a standard Transformer;
\item
  a parameter-matched Transformer;
\item
  a phase-only model;
\item
  a wavelength-hierarchy-only model;
\item
  a direction-only model;
\item
  a geodesic-distance-only model;
\item
  full WavePhase Attention;
\item
  no spectral selection, a classical hard mask, a Chebyshev filter, and
  quantum simulation;
\item
  with and without RAG; and
\item
  covariant phase, naive phase difference, and a phase-shuffled control.
\end{enumerate}

\subsection{16.4 Evaluation Metrics}\label{evaluation-metrics}

\textbf{Language-model performance:} perplexity, exact match, F1, and
long-form quality.

\textbf{Hierarchy:} Spearman correlation, HyperLex correlation,
hypernym-direction accuracy, inclusion AUC, transitivity consistency,
and closure under lattice operations.

\textbf{Directional stability:} alignment with summary embeddings, mean
length before topic drift, paragraph-level instruction consistency, and
expert evaluation.

\textbf{Factuality:} atomic factual precision, FEVER score, TruthfulQA
truthfulness, and SAFE/FActScore.

\textbf{Calibration:} Brier score, negative log-likelihood, ECE, AUROC,
AUPRC, risk--coverage, and answer rate at a specified error rate.

\textbf{Efficiency:} training time, inference latency, GPU memory, graph
eigensolver time, quantum gate count, circuit depth, shots, and
performance under noise.

\subsection{16.5 Statistical Design}\label{statistical-design}

\begin{itemize}
\tightlist
\item
  Use at least five random seeds and report means and 95\% confidence
  intervals.
\item
  Use paired bootstrap or an appropriate paired test on the same inputs.
\item
  Match the hyperparameter-search budget across baselines.
\item
  Separate calibration data from the final test set.
\item
  Predefine primary metrics and rejection criteria.
\item
  Report effect size and computational cost together; do not conclude
  from statistical significance alone.
\end{itemize}

\subsection{16.6 Falsification Criteria}\label{falsification-criteria}

Any of the following results would require rejection or revision of a
central hypothesis:

\begin{enumerate}
\def\labelenumi{\arabic{enumi}.}
\tightlist
\item
  learned wavelengths do not consistently correlate with known concept
  hierarchies;
\item
  a multi-axis model still fails to outperform Order or Poincaré
  embeddings in order prediction;
\item
  the low-frequency direction increases topic drift relative to summary-
  or instruction-based directions;
\item
  the covariant phase term cannot be distinguished from a phase-shuffled
  control;
\item
  geometric risk does not outperform entropy, semantic entropy, or a
  standalone evidence verifier;
\item
  curvature cannot distinguish legitimate topic transitions from errors;
  or
\item
  neither simulation nor hardware shows an end-to-end advantage after
  including noise, state preparation, and readout.
\end{enumerate}

\section{17. Prototype Experiments with the Validation
Studio}\label{prototype-experiments-with-the-validation-studio}

\subsection{17.1 Purpose and Implementation
Scope}\label{purpose-and-implementation-scope}

Validation Studio v1.0.0 is a fully offline minimal implementation that
visualizes the classical quantum-inspired realization of Section 14.1 in
the following order: text input, semantic graph, low-frequency discourse
direction, covariant phase and semantic wavelength, self-adjoint
spectral selection, amplitude reweighting, WavePhase Attention, and
evidence/risk evaluation. The same interface recomputes the complexity
comparison of Section 15.1 and RQ1--RQ5 of Section 16.1 under one
configuration and can record the results as JSON or PDF.

The numerical results in this section reproduce the built-in synthetic
example documented in the user manual. They are not final results on
public external benchmarks or independently collected data. They should
be interpreted as prototype evidence that the theoretical mechanisms,
evaluation metrics, and falsification criteria are connected
consistently in software.

\begin{figure}
\centering
\includegraphics[width=6.3in,height=\textheight]{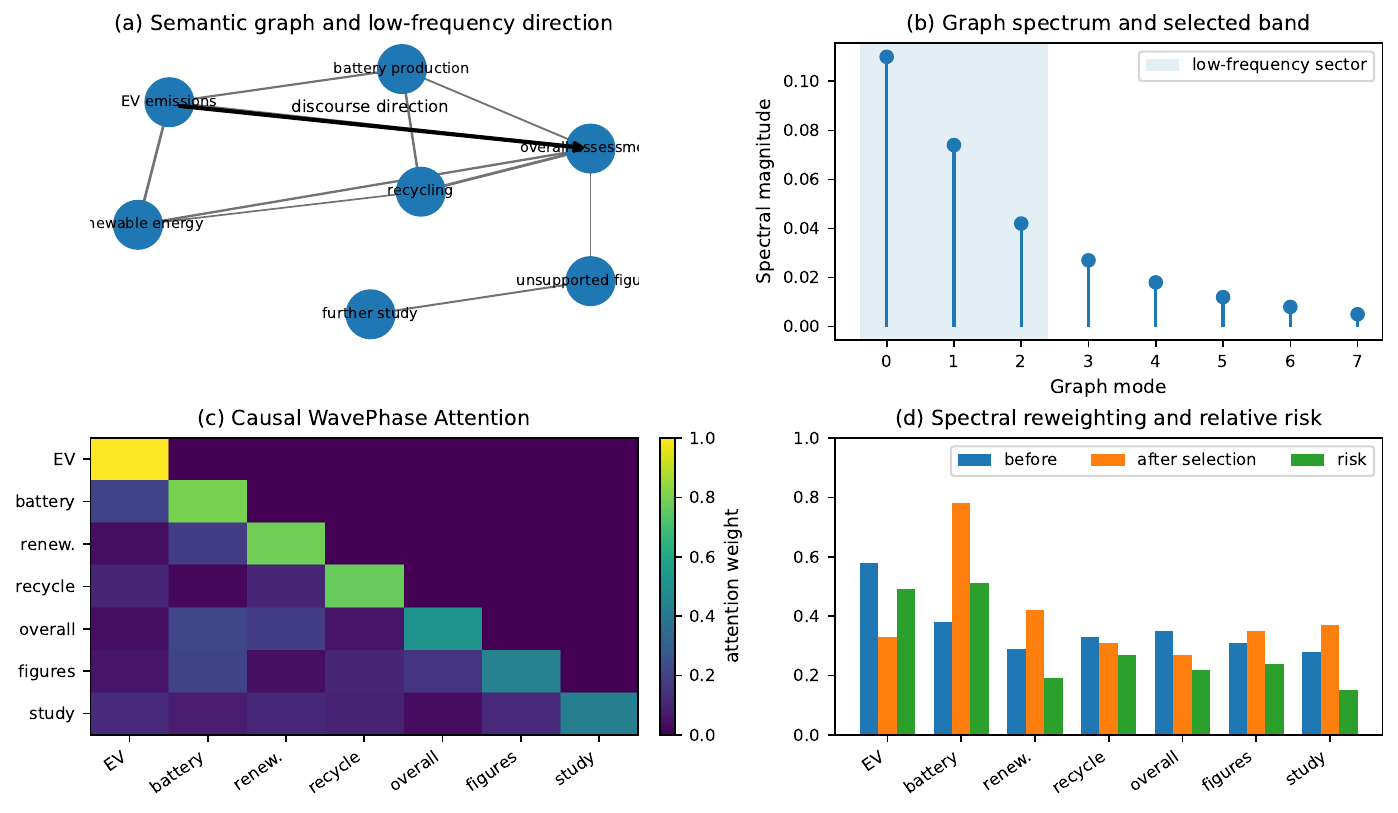}
\caption{Visualization of the semantic graph, spectrum, WavePhase
Attention, and reweighting/risk in the Validation Studio.}
\end{figure}

\subsection{17.2 Experimental Conditions and Reproduction
Procedure}\label{experimental-conditions-and-reproduction-procedure}

We fixed the random seed at 20260720, used 240 samples, observation
noise 0.22, and circuit noise 0.08, and repeated the experiment over
five seeds before recomputing all research questions. Every RQ used the
same built-in synthetic generator and decision rules. We evaluated
differences from the baseline, the sign of the 95\% interval,
discrimination and calibration metrics, and computational resources.

\begin{longtable}[]{@{}
  >{\raggedright\arraybackslash}p{(\columnwidth - 4\tabcolsep) * \real{0.2927}}
  >{\raggedright\arraybackslash}p{(\columnwidth - 4\tabcolsep) * \real{0.2927}}
  >{\raggedright\arraybackslash}p{(\columnwidth - 4\tabcolsep) * \real{0.3902}}@{}}
\toprule\noalign{}
\begin{minipage}[b]{\linewidth}\raggedright
\textbf{Item}
\end{minipage} & \begin{minipage}[b]{\linewidth}\raggedright
\textbf{Setting}
\end{minipage} & \begin{minipage}[b]{\linewidth}\raggedright
\textbf{Role}
\end{minipage} \\
\midrule\noalign{}
\endhead
\bottomrule\noalign{}
\endlastfoot
Random seed & 20260720 & Reproduction of built-in synthetic data \\
Sample size & 240 & Evaluation samples for each RQ \\
Observation noise & 0.22 & Measurement variation in RQ1--RQ4 \\
Circuit noise & 0.08 & Degradation in the RQ5 circuit simulation \\
Number of seeds & 5 & Repetitions for means and intervals \\
\end{longtable}

Table 1. Experimental conditions for the built-in synthetic example.

\subsection{17.3 Overall Results}\label{overall-results}

Of the five research questions, RQ1--RQ4 met the prespecified support
criteria, whereas RQ5 became a falsification candidate for the
end-to-end cost efficiency of the quantum-executable realization. The
experiment therefore provides initial support for the main classical
quantum-inspired components, but not for quantum computational
advantage. The mean evidence-strength value of approximately 73\% is
only a relative visualization index within the Studio; it is neither the
probability that the paper is correct nor a probability of truth.

\begin{figure}
\centering
\includegraphics[width=6in,height=\textheight]{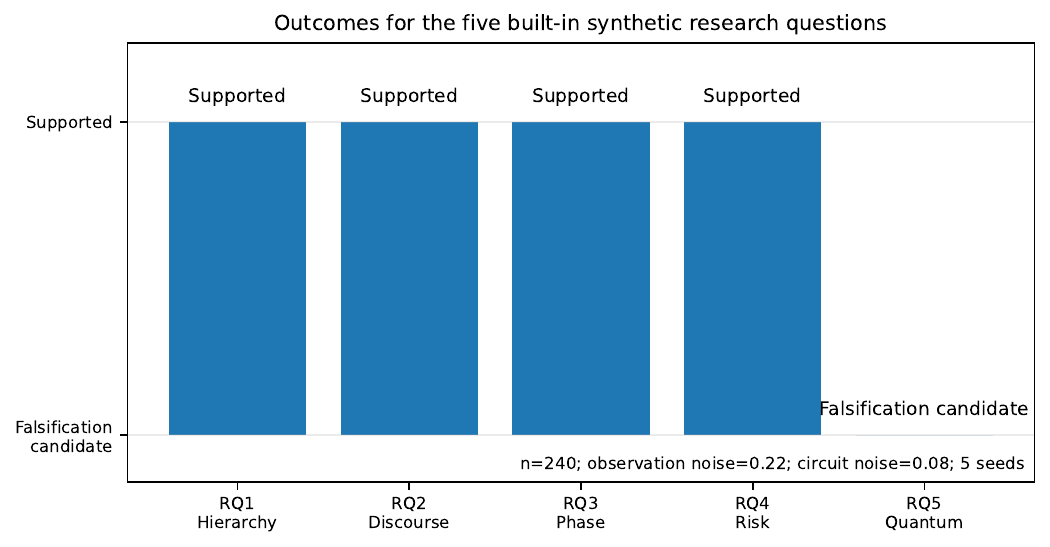}
\caption{Outcomes for the five research questions in the built-in
synthetic example.}
\end{figure}

\begin{longtable}[]{@{}
  >{\raggedright\arraybackslash}p{(\columnwidth - 8\tabcolsep) * \real{0.1531}}
  >{\raggedright\arraybackslash}p{(\columnwidth - 8\tabcolsep) * \real{0.2347}}
  >{\raggedright\arraybackslash}p{(\columnwidth - 8\tabcolsep) * \real{0.2449}}
  >{\raggedright\arraybackslash}p{(\columnwidth - 8\tabcolsep) * \real{0.1939}}
  >{\raggedright\arraybackslash}p{(\columnwidth - 8\tabcolsep) * \real{0.1531}}@{}}
\toprule\noalign{}
\begin{minipage}[b]{\linewidth}\raggedright
\textbf{RQ}
\end{minipage} & \begin{minipage}[b]{\linewidth}\raggedright
\textbf{Primary comparison}
\end{minipage} & \begin{minipage}[b]{\linewidth}\raggedright
\textbf{QuantumPhaseNet}
\end{minipage} & \begin{minipage}[b]{\linewidth}\raggedright
\textbf{Baseline}
\end{minipage} & \begin{minipage}[b]{\linewidth}\raggedright
\textbf{Outcome}
\end{minipage} \\
\midrule\noalign{}
\endhead
\bottomrule\noalign{}
\endlastfoot
RQ1 hierarchy & Spearman correlation & 0.852 & 0.707 & Supported \\
RQ2 discourse & Mean discourse alignment & 0.933 & 0.589 & Supported \\
RQ3 phase & AUROC & 0.881 & 0.765 (shuffle 0.536) & Supported \\
RQ4 risk & Error-detection AUROC & 0.854 & entropy 0.634 & Supported \\
RQ5 quantum & End-to-end efficiency & quantum sim. 0.107 & Chebyshev
0.707 & Falsification candidate \\
\end{longtable}

Table 2. Main results for RQ1--RQ5.

\subsection{17.4 RQ1: Semantic Wavelength and Concept
Hierarchy}\label{rq1-semantic-wavelength-and-concept-hierarchy}

The rank correlation between semantic wavelength and known abstraction
depth was 0.852, exceeding the simple embedding-scale baseline of 0.707
by 0.145. Hypernym-direction accuracy was 87.3\%, and inclusion AUC was
0.953. In the built-in example, inverse-frequency wavelength therefore
acted as a plausible one-dimensional proxy for conceptual hierarchy.

\begin{figure}
\centering
\includegraphics[width=6.3in,height=\textheight]{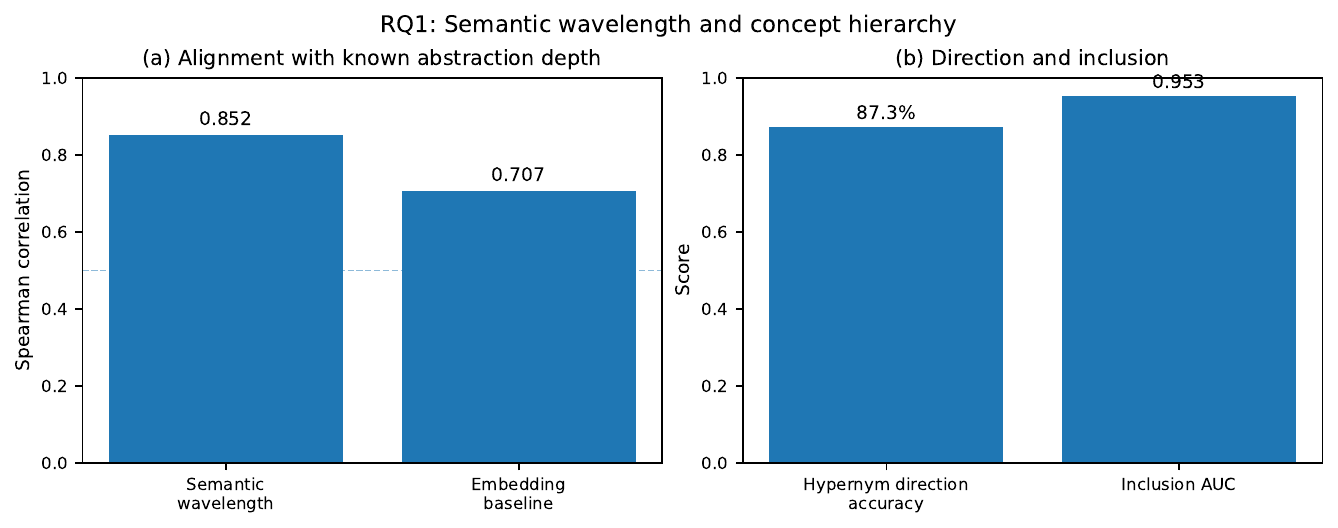}
\caption{RQ1: semantic-wavelength scale and concept-hierarchy metrics.}
\end{figure}

\subsection{17.5 RQ2: Low-Frequency Discourse Direction and Topic
Retention}\label{rq2-low-frequency-discourse-direction-and-topic-retention}

Mean discourse alignment with the low-frequency direction was 0.933,
compared with 0.589 for the baseline. Mean paragraphs before drift were
41.2 versus 16.2, and the aligned-paragraph rate was 95.4\% versus
56.3\%; the lower bound of the across-seed 95\% interval was 0.061.
These results suggest that a direction vector constructed from
low-frequency components may help retain the intended direction in
long-form generation.

\begin{figure}
\centering
\includegraphics[width=6.3in,height=\textheight]{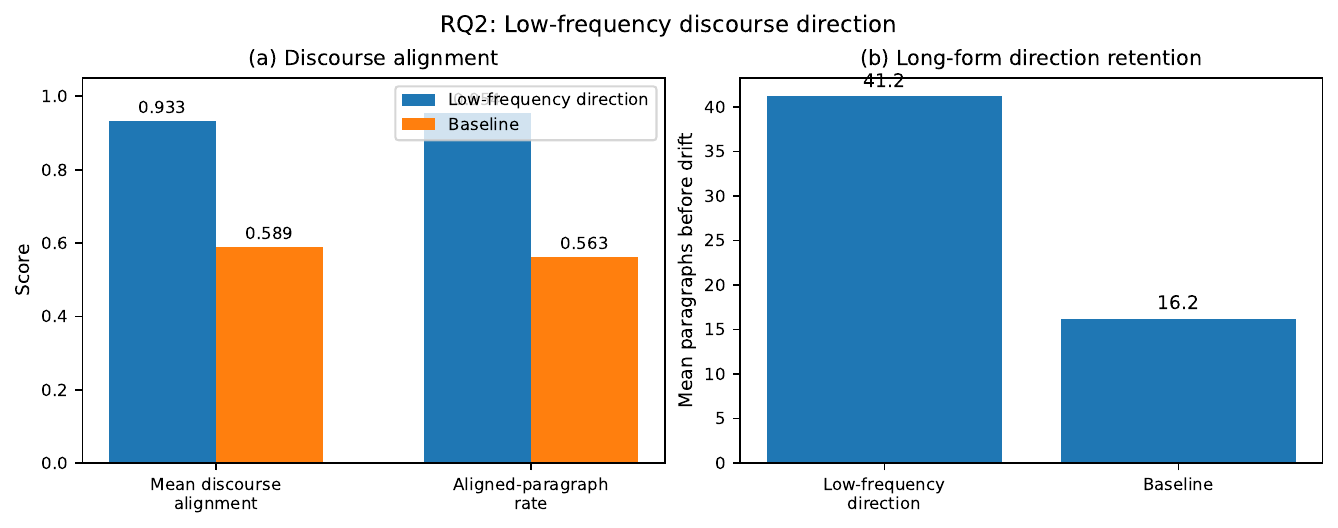}
\caption{RQ2: discourse alignment and suppression of drift by the
low-frequency direction.}
\end{figure}

\subsection{17.6 RQ3: Predictive Power of Covariant Phase
Coherence}\label{rq3-predictive-power-of-covariant-phase-coherence}

Covariant phase coherence achieved AUROC 0.881, exceeding cosine
similarity at 0.765 and the phase-shuffled control at 0.536.
Classification accuracy was 81.5\% versus 69.2\%, and the Brier score
was 0.154 versus 0.197. The decline of the shuffled control toward
chance suggests that the connection-aware phase structure, rather than
merely additional parameters, may carry predictive information.

\begin{figure}
\centering
\includegraphics[width=6.3in,height=\textheight]{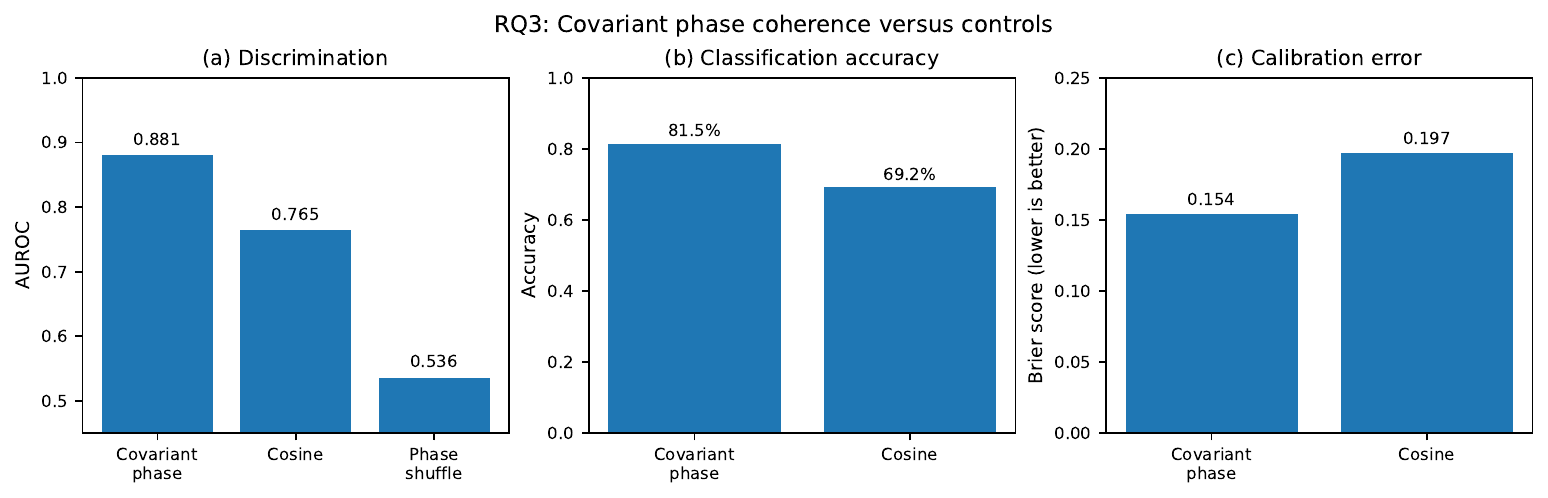}
\caption{RQ3: comparison of covariant phase, cosine similarity, and a
phase-shuffled control.}
\end{figure}

\subsection{17.7 RQ4: Geometric Risk and
Hallucination}\label{rq4-geometric-risk-and-hallucination}

The error-detection AUROC of geometric risk was 0.854, compared with
0.634 for the entropy baseline. Calibration metrics were Brier 0.150 and
ECE 0.098. At an allowed error rate of 10\%, the answer rate was 53.1\%,
compared with 3.9\% for entropy. A risk integrating directional
deviation, evidence deficiency, curvature, and uncertainty may therefore
improve risk--coverage tradeoffs in selective generation.

\begin{figure}
\centering
\includegraphics[width=6.3in,height=\textheight]{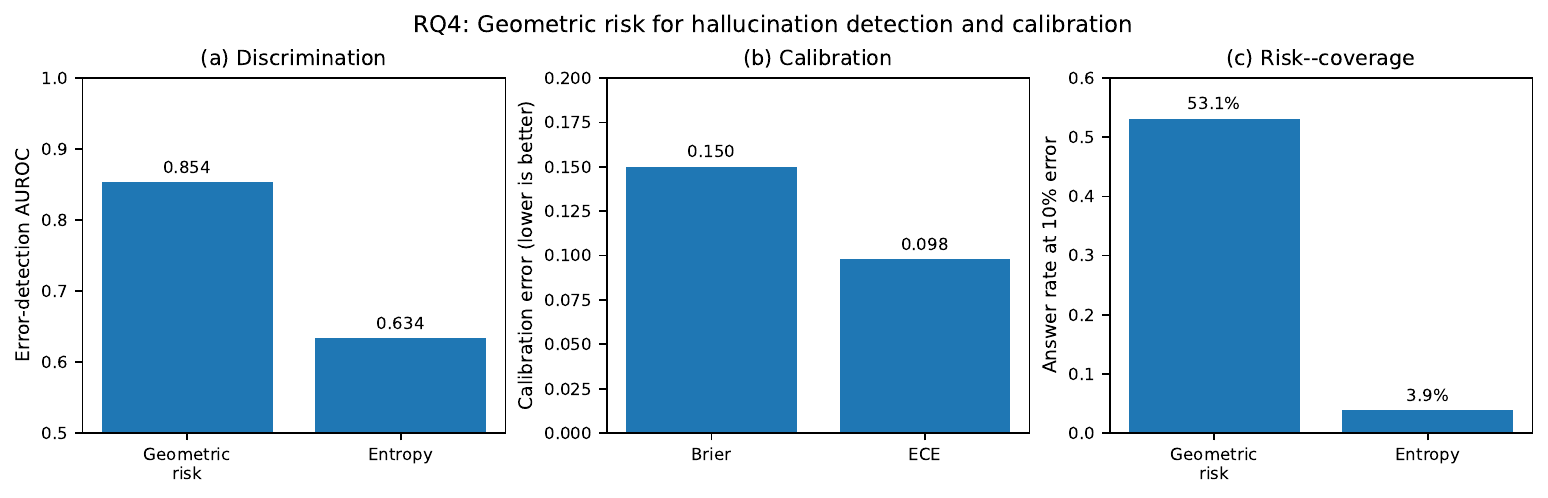}
\caption{RQ4: discrimination, calibration, and risk--coverage of
geometric risk.}
\end{figure}

\subsection{17.8 RQ5: Quantum Selection and Classical
Approximation}\label{rq5-quantum-selection-and-classical-approximation}

At circuit noise 0.08, the target-subspace probability in quantum
simulation was 25.5\%, below the 70.7\% achieved by the classical
Chebyshev approximation. End-to-end cost efficiency, including state
preparation, circuit depth, shots, and readout, was 0.107 versus 0.707.
Thus, the quantum-executable realization showed no advantage under this
configuration. This result does not invalidate the classical
quantum-inspired realization and is consistent with the deliberately
limited claim about quantum advantage in Section 15.2.

\begin{figure}
\centering
\includegraphics[width=6.3in,height=\textheight]{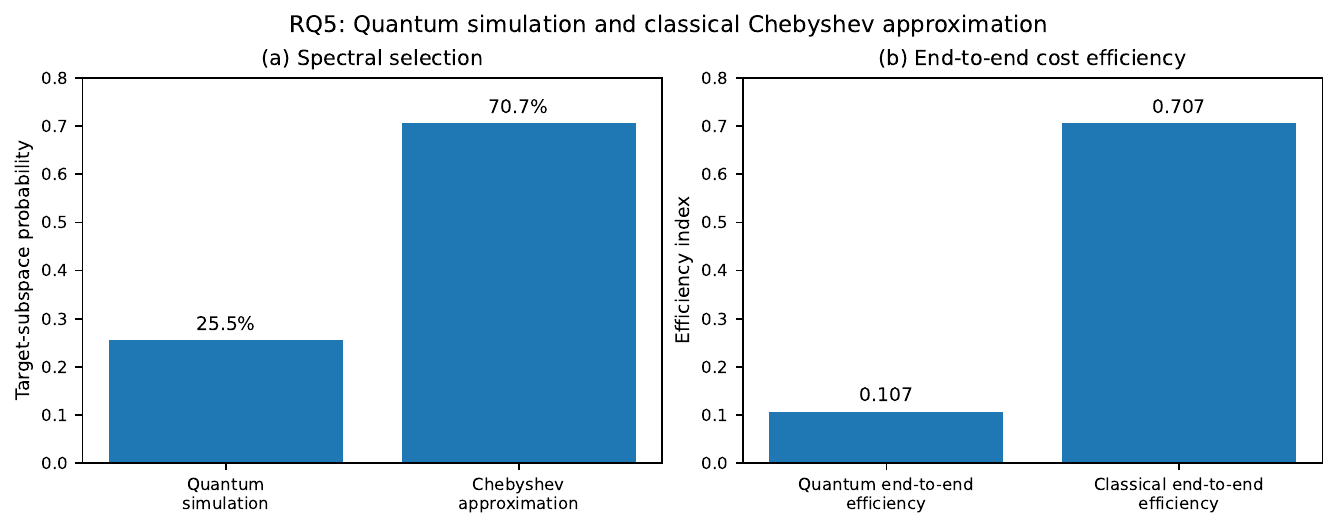}
\caption{RQ5: quantum-circuit simulation versus the classical Chebyshev
approximation.}
\end{figure}

\subsection{17.9 Complexity Validation for Section
15.1}\label{complexity-validation-for-section-15.1}

The Studio compares theoretical curves and browser measurements for
dense attention and all-pairs WavePhase, both \(O(n^2d)\); full
eigendecomposition, \(O(n^3)\); sparse Lanczos, approximately
\(O(TK|E|)\); and a Chebyshev filter, linear in the number of edges,
polynomial degree, and feature dimension. Figure 8 is a schematic
visualization of these asymptotic expressions under fixed representative
parameters, not a plot of measured runtimes. Preserving the advantage of
sparse approximation requires avoiding exact all-pairs \(k\)NN and
accounting for preprocessing such as ANN, Nystrom, landmarks, and graph
reuse.

\begin{figure}
\centering
\includegraphics[width=6.1in,height=\textheight]{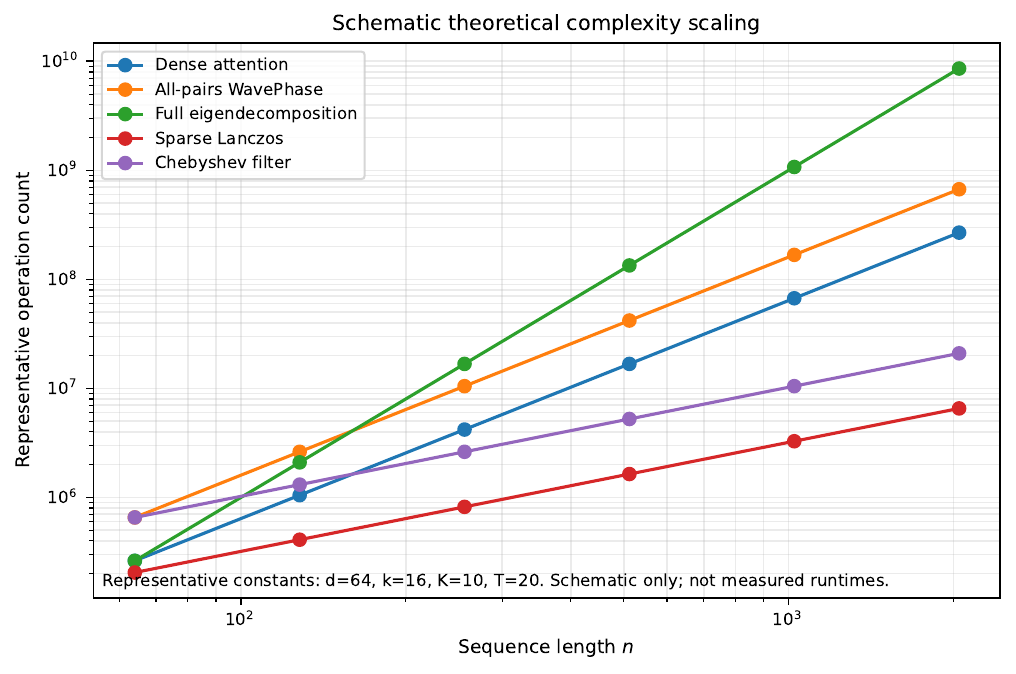}
\caption{Schematic theoretical complexity scaling from Section 15.1.}
\end{figure}

\subsection{17.10 Interpretation and Scope of
Validity}\label{interpretation-and-scope-of-validity}

The improvements in RQ1--RQ4 show that wavelength, low-frequency
direction, covariant phase, and geometric risk produce baseline
differences at least under the designed synthetic conditions. This
strengthens the internal consistency and implementability of the
mechanisms, but it does not replace external validation on independent
datasets such as WordNet, HyperLex, TruthfulQA, FEVER, and SAFE. Future
experiments should separate training, calibration, and final testing;
use paired comparisons on identical inputs; run at least ten seeds;
preregister primary metrics; and report effect sizes and computational
cost together.

For scientific completeness, the unsupported result for RQ5 is retained.
Removing it would erase the distinction between the quantum-executable
and classical quantum-inspired realizations and would undermine the
falsifiability built into the theory from the outset.

\section{18. Identifiability, Limitations, and Ethical
Considerations}\label{identifiability-limitations-and-ethical-considerations}

\subsection{18.1 Nonuniqueness of Gauges and
Coordinates}\label{nonuniqueness-of-gauges-and-coordinates}

The embedding map \(\Phi\), metric \(g\), connection \(\nabla\), and
local frames are not unique. Observable quantities are invariants such
as geodesic distances, inner products after parallel transport, Wilson
loops, and covariant phase rates. The phase angle or connection
coefficients themselves should not be interpreted as intrinsic semantic
entities.

\subsection{18.2 Nonuniqueness of
Hierarchy}\label{nonuniqueness-of-hierarchy}

The same pair of concepts may participate in different hierarchical
axes, including type, part--whole structure, timescale, causal
abstraction, and social category. A single wavelength cannot be treated
as a universal concept depth. Multi-axis scales and supervised relation
types are required.

\subsection{18.3 Low-Frequency Bias}\label{low-frequency-bias}

Misinformation repeated over a long span may itself become a
low-frequency component. Discourse direction does not guarantee truth
and must be evaluated independently from evidence verification.

\subsection{18.4 Limits of Hallucination
Risk}\label{limits-of-hallucination-risk}

Truth labels are difficult to define for consistently wrong generations,
incorrect retrieval sources, value judgments, future predictions, and
ambiguous questions. Risk scores depend on the use case and annotation
protocol. In high-stakes domains, generated text should not serve as the
sole basis for decisions; human review and source presentation should be
incorporated.

\subsection{18.5 Quantum Advantage Is Not
Established}\label{quantum-advantage-is-not-established}

The partial complexity of QFT or amplitude amplification does not
establish an end-to-end speedup. Classical alternatives include FFT,
sparse-matrix methods, Lanczos, Chebyshev approximation, and GPU
parallelization. A fair comparison must match input and output models
and target precision.

\section{19. Minimal Reproducible
Model}\label{minimal-reproducible-model}

In the first implementation stage, set

\[\mathcal{M} = \mathbb{R}^{d},\quad\quad g = I,\quad\quad\nabla = d,\quad\quad F_{\nabla} = 0,\quad\quad c_{s} = 1\]

Then geodesic distance reduces to Euclidean distance, parallel transport
is the identity, and covariant phase coherence becomes the ordinary
complex inner product. We recommend the following staged implementation.

\begin{enumerate}
\def\labelenumi{\arabic{enumi}.}
\tightlist
\item
  \textbf{Phase only:} complex projection and phase coherence.
\item
  \textbf{Wavelength hierarchy:} \(\Omega\), \(\lambda\), and
  HyperLex/WordNet losses.
\item
  \textbf{Direction:} graph Laplacian and \(\Theta_S\).
\item
  \textbf{WavePhase Attention:} combine phase, wavelength, direction,
  and distance.
\item
  \textbf{Evidence risk:} RAG, claim decomposition, calibration, and
  abstention.
\item
  \textbf{Learned geometry:} \(G_\theta\), a nontrivial connection, and
  curvature.
\item
  \textbf{Quantum simulation:} small circuits for QPE, the oracle, and
  amplitude amplification.
\end{enumerate}

At each stage, verify that the contribution from the preceding stage is
reproducible before introducing all mechanisms at once. This makes it
easier to identify which component contributes to performance,
instability, or computational cost.

\section{20. Correspondence Between Main Results and Empirical
Hypotheses}\label{correspondence-between-main-results-and-empirical-hypotheses}

\begin{longtable}[]{@{}
  >{\raggedright\arraybackslash}p{(\columnwidth - 4\tabcolsep) * \real{0.1633}}
  >{\raggedright\arraybackslash}p{(\columnwidth - 4\tabcolsep) * \real{0.1633}}
  >{\raggedright\arraybackslash}p{(\columnwidth - 4\tabcolsep) * \real{0.6599}}@{}}
\toprule\noalign{}
\begin{minipage}[b]{\linewidth}\raggedright
\textbf{Category}
\end{minipage} & \begin{minipage}[b]{\linewidth}\raggedright
\textbf{Result}
\end{minipage} & \begin{minipage}[b]{\linewidth}\raggedright
\textbf{Content}
\end{minipage} \\
\midrule\noalign{}
\endhead
\bottomrule\noalign{}
\endlastfoot
Theorem & Covariant phase rate & \(\Omega_\gamma\) and
\(\lambda_\gamma\) are invariant under local gauge transformations \\
Theorem & Quotient hierarchy & Single-wavelength equivalence classes are
totally ordered and form a distributive lattice under binary
operations \\
Proposition & Multi-axis hierarchy & A distributive lattice results when
the image is closed under componentwise min/max \\
Theorem & Oracle & \(O_G\) is self-adjoint and involutory \\
Theorem & Amplitude amplification & Success probability is
\(\sin^2((2k+1)\theta)\) \\
Theorem & Quantum block & The full block is unitary when its components
are unitary \\
Theorem & WavePhase invariance & Logits built from invariant features
are locally gauge invariant \\
Theorem & Attention boundedness & The output remains in the convex hull
of the value vectors \\
Theorem & Local stability & Attention is Lipschitz under bounded
Lipschitz features \\
Conditional theorem & Generation convergence & Linear convergence under
Hadamard, strong-convexity, and smoothness assumptions \\
Empirical hypothesis & Long wavelength = higher concept & Built-in
synthetic result: Spearman 0.852 (baseline 0.707), direction accuracy
87.3\%, inclusion AUC 0.953 \\
Empirical hypothesis & Low frequency = discourse & Built-in synthetic
result: mean alignment 0.933 (baseline 0.589), aligned-paragraph rate
95.4\% (baseline 56.3\%) \\
Empirical hypothesis & Risk predicts error & Built-in synthetic result:
AUROC 0.854 (entropy 0.634), Brier 0.150, ECE 0.098 \\
Empirical hypothesis & Predictive covariant phase & Built-in synthetic
result: AUROC 0.881 (cosine 0.765, shuffled phase 0.536), Brier 0.154 \\
Not established & Quantum advantage & Built-in synthetic result: quantum
end-to-end efficiency 0.107 versus classical efficiency 0.707; no
quantum advantage observed \\
\end{longtable}

\section{21. Conclusion}\label{conclusion}

This paper models meaning through

\[\mathfrak S=(\mathcal M,g,\mu,E,\nabla,\mathcal H,H_{\mathrm{sem}},\boldsymbol\lambda,\Theta_S,\mathcal R)\]

In the built-in synthetic example of the Validation Studio, the RQ1
hierarchy scale, RQ2 low-frequency discourse direction, RQ3 covariant
phase coherence, and RQ4 geometric risk each outperformed their
respective baselines. These results provide initial support for the
implementability of the classical quantum-inspired realization and for
separating its mechanisms experimentally. In particular, hierarchy
correlation 0.852, discourse alignment 0.933, covariant-phase AUROC
0.881, and geometric-risk AUROC 0.854 demonstrate that the proposed
invariants and spectral direction can be connected to measurable
evaluation criteria.

At the same time, the RQ5 quantum-circuit simulation underperformed the
classical Chebyshev approximation in end-to-end cost, so quantum
advantage was not supported. This negative result strengthens the
paper's distinction between the quantum-executable and classical
quantum-inspired realizations. The defensible empirical conclusion at
this stage is therefore limited: the main QuantumPhaseNet mechanisms
based on classical linear algebra, sparse graphs, and polynomial filters
are testable under synthetic conditions.

The central next steps are (i) independent replication of the hierarchy
results on WordNet and HyperLex, (ii) evaluation of discourse retention
on long-form and Japanese data, (iii) calibration of risk on TruthfulQA,
FEVER, and SAFE, (iv) discrimination between curvature caused by errors
and curvature caused by legitimate topic transitions, and (v) a fair
classical--quantum resource comparison from state preparation through
readout.

\section{Appendix A. Notation}\label{appendix-a.-notation}

\begin{longtable}[]{@{}ll@{}}
\toprule\noalign{}
Symbol & Meaning \\
\midrule\noalign{}
\endhead
\bottomrule\noalign{}
\endlastfoot
\(\mathcal{M}\) & semantic concept manifold \\
\(g\) & Riemannian metric \\
\(\mu\) & measure on the semantic manifold \\
\(E\rightarrow\mathcal{M}\) & complex semantic vector bundle \\
\(\nabla\) & unitary connection \\
\(F_{\nabla}\) & connection curvature \\
\(\Psi\in\mathcal{H}\) & normalized semantic state \\
\(\Omega_{\gamma}\) & covariant phase rate \\
\(\ell_{\varphi,\gamma}\) & accumulated covariant phase \\
\(\lambda\) & semantic wavelength / proxy for conceptual scale \\
\(C_{ij}^{\nabla}\) & phase coherence including parallel transport \\
\(L\) & graph Laplacian \\
\(L_{\nabla}\) & connection Laplacian \\
\(\Theta_S\) & discourse-direction vector \\
\(H_{\mathrm{sem}}\) & self-adjoint semantic Hamiltonian \\
\(F_N\) & external \(N\)-dimensional QFT/DFT \\
\(\Pi_G\) & projector onto the target semantic subspace \\
\(O_G\) & phase-flip oracle \\
\(Q\) & amplitude-amplification iteration \\
\(\ell_{ij}^{\mathrm{WP}}\) & WavePhase logit \\
\(\mathcal{R}_t\) & stepwise hallucination risk \\
\end{longtable}

\section{Appendix B. Invariance Tests for
Implementation}\label{appendix-b.-invariance-tests-for-implementation}

The training code should include the following unit tests:

\begin{enumerate}
\def\labelenumi{\arabic{enumi}.}
\tightlist
\item
  Add random local phases \(\chi_i\) and transform the connection
  consistently; verify that \(\Omega_i\), \(C_{ij}^{\nabla}\), the
  logits, and the output are invariant.
\item
  Change eigenvector signs, phases, and the basis within degenerate
  eigenspaces; verify that spectral projectors are invariant.
\item
  Change the padding length \(N\) and verify that the output on
  nonpadding positions agrees within tolerance.
\item
  In the limiting cases in which every state or no state is marked,
  verify that amplitude amplification degenerates as expected.
\item
  Verify recovery of standard attention when
  \(\alpha_\varphi=\alpha_\lambda=\alpha_d=\alpha_g=\alpha_e=0\).
\item
  Verify that phase-shuffled, wavelength-shuffled, and
  direction-shuffled controls perform consistently worse than the active
  model.
\end{enumerate}

\section{References}\label{references}

Abbasi-Yadkori, Yasin, Ilja Kuzborskij, David Stutz, et al.~2024.
``Mitigating LLM Hallucinations via Conformal Abstention.''
arXiv:2405.01563.

Angelopoulos, Anastasios N., Stephen Bates, Adam Fisch, Lihua Lei, and
Tal Schuster. 2022. ``Conformal Risk Control.'' arXiv:2208.02814.

Biamonte, Jacob, Peter Wittek, Nicola Pancotti, Patrick Rebentrost,
Nathan Wiebe, and Seth Lloyd. 2017. ``Quantum Machine Learning.''
\emph{Nature} 549: 195--202. arXiv:1611.09347.

Bronstein, Michael M., Joan Bruna, Taco Cohen, and Petar Veličković.
2021. ``Geometric Deep Learning: Grids, Groups, Graphs, Geodesics, and
Gauges.'' arXiv:2104.13478.

Coecke, Bob, Giovanni de Felice, Konstantinos Meichanetzidis, and Alexis
Toumi. 2020. ``Foundations for Near-Term Quantum Natural Language
Processing.'' arXiv:2012.03755.

Coecke, Bob, Mehrnoosh Sadrzadeh, and Stephen Clark. 2010.
``Mathematical Foundations for a Compositional Distributional Model of
Meaning.'' \emph{Linguistic Analysis} 36 (1--4): 345--384.
arXiv:1003.4394.

Coppersmith, Don. 1994. ``An Approximate Fourier Transform Useful in
Quantum Factoring.'' arXiv:quant-ph/0201067.

Farquhar, Sebastian, Jannik Kossen, Lorenz Kuhn, and Yarin Gal. 2024.
``Detecting Hallucinations in Large Language Models Using Semantic
Entropy.'' \emph{Nature} 630: 625--630. doi:10.1038/s41586-024-07421-0.

Fukiya, Kazuo, and Kiyotaka Kasubuchi. 2026. ``QuantumPhaseNet
Validation Studio v1.0.0: User Manual with Worked Examples.'' July 27,
2026.

Gilyén, András, Yuan Su, Guang Hao Low, and Nathan Wiebe. 2019.
``Quantum Singular Value Transformation and Beyond: Exponential
Improvements for Quantum Matrix Arithmetics.'' \emph{Proceedings of STOC
2019}. arXiv:1806.01838.

Grover, Lov K. 1996. ``A Fast Quantum Mechanical Algorithm for Database
Search.'' \emph{Proceedings of STOC 1996}, 212--219.
arXiv:quant-ph/9605043.

Kasubuchi, Kiyotaka, and Kazuo Fukiya. 2026. ``WavePhaseNet: A DFT-Based
Method for Constructing Semantic Conceptual Hierarchy Structures
(SCHS).'' arXiv:2602.14419.

Kitaev, Alexei Yu. 1995. ``Quantum Measurements and the Abelian
Stabilizer Problem.'' arXiv:quant-ph/9511026.

Lewis, Patrick, Ethan Perez, Aleksandra Piktus, et al.~2020.
``Retrieval-Augmented Generation for Knowledge-Intensive NLP Tasks.''
\emph{NeurIPS} 33: 9459--9474. arXiv:2005.11401.

Lin, Stephanie, Jacob Hilton, and Owain Evans. 2022. ``TruthfulQA:
Measuring How Models Mimic Human Falsehoods.'' \emph{ACL 2022},
3214--3252. arXiv:2109.07958.

Manakul, Potsawee, Adian Liusie, and Mark J. F. Gales. 2023.
``SelfCheckGPT: Zero-Resource Black-Box Hallucination Detection for
Generative Large Language Models.'' \emph{EMNLP 2023}, 9004--9017.
arXiv:2303.08896.

Min, Sewon, Kalpesh Krishna, Xinxi Lyu, et al.~2023. ``FActScore:
Fine-Grained Atomic Evaluation of Factual Precision in Long Form Text
Generation.'' \emph{EMNLP 2023}. arXiv:2305.14251.

Nickel, Maximilian, and Douwe Kiela. 2017. ``Poincaré Embeddings for
Learning Hierarchical Representations.'' \emph{NeurIPS 2017}.
arXiv:1705.08039.

Nielsen, Michael A., and Isaac L. Chuang. 2010. \emph{Quantum
Computation and Quantum Information}. 10th Anniversary ed.~Cambridge
University Press.

Schuld, Maria, and Nathan Killoran. 2019. ``Quantum Machine Learning in
Feature Hilbert Spaces.'' \emph{Physical Review Letters} 122: 040504.
arXiv:1803.07128.

Shuman, David I., Benjamin Ricaud, and Pierre Vandergheynst. 2013.
``Vertex-Frequency Analysis on Graphs.'' arXiv:1307.5708.

Singer, Amit, and Hau-tieng Wu. 2012. ``Vector Diffusion Maps and the
Connection Laplacian.'' \emph{Communications on Pure and Applied
Mathematics} 65 (8): 1067--1144. arXiv:1102.0075.

Thorne, James, Andreas Vlachos, Christos Christodoulopoulos, and Arpit
Mittal. 2018. ``FEVER: A Large-Scale Dataset for Fact Extraction and
VERification.'' \emph{NAACL 2018}. arXiv:1803.05355.

Vaswani, Ashish, Noam Shazeer, Niki Parmar, et al.~2017. ``Attention Is
All You Need.'' \emph{NeurIPS} 30. arXiv:1706.03762.

Vendrov, Ivan, Ryan Kiros, Sanja Fidler, and Raquel Urtasun. 2016.
``Order-Embeddings of Images and Language.'' \emph{ICLR 2016}.
arXiv:1511.06361.

Vulić, Ivan, Daniela Gerz, Douwe Kiela, Felix Hill, and Anna Korhonen.
2016. ``HyperLex: A Large-Scale Evaluation of Graded Lexical
Entailment.'' \emph{Computational Linguistics} 43 (4): 781--835.
arXiv:1608.02117.

Wei, Jerry, Chengrun Yang, Xinying Song, et al.~2024. ``Long-Form
Factuality in Large Language Models.'' arXiv:2403.18802.

\end{document}